\documentclass{article}

\PassOptionsToPackage{authoryear,round}{natbib}
\usepackage[preprint]{neurips_2026}

\usepackage[utf8]{inputenc}
\usepackage[T1]{fontenc}
\usepackage{amsmath,amssymb}
\usepackage{booktabs}
\usepackage{graphicx}
\usepackage{nicefrac}
\usepackage{microtype}
\usepackage{xcolor}
\usepackage{url}
\usepackage{hyperref}
\hypersetup{
  hidelinks,
  pdftitle={A Step Towards Forgetting: Optimiser History and the Loss of Answer Mass},
  pdfauthor={Valeria Ruscio, Seth Nabarro, Keiran Thompson}
}

\newcommand{\dd}{\mathrm{d}}

\title{A Step Towards Forgetting:\\Optimiser History and the Loss of Answer Mass}
\author{%
  Valeria Ruscio \quad Seth Nabarro \quad Keiran Thompson \\
  Intuition Machines Inc
}

\begin{document}

\maketitle

\begin{abstract}
During fine-tuning, a language model can assign less probability to previously learned answers even when the current gradient acts to preserve that probability. With momentum, each update also carries gradients computed at earlier model states, and these stored contributions can push the model in the opposite direction. We investigate how this optimiser memory contributes to forgetting by separating old-task loss into confusion among its answers and leakage of probability outside the answer set. Across three language-model families, answer mass consistently declines while discrimination among old answers usually improves: the model becomes less likely to produce answers that it can still distinguish correctly.\\
Decomposing Adam updates shows opposing contributions to this loss of answer mass. Over training, accumulated history favours leakage, while the current gradient opposes it. Resolving history by age shows that the harmful contributions come mainly from older gradients of the new task, whereas recent gradients tend to protect the old answers. Changes in history’s effect are dominated by its orientation relative to the old-task gradient. Interventions that reset momentum while matching the initial update norm establish that stored history affects retention, with state-dependent immediate effects and lower final old-task loss over longer Adam continuations, mainly through recovered answer mass. Finally, integration along finite updates shows that most sampled large loss increases are captured by local projections, while curvature along history amplifies some events. Together, these findings show how an optimiser’s memory can erode learned behaviour even as its current gradient acts to preserve it.
\end{abstract}

\section{Introduction}
When a language model is trained on a new task, it may stop producing answers that it previously gave correctly, and this failure can arise through two different changes in its output distribution, since probability can move between the old answers, weakening the model's ability to discriminate among them, or it can leave the old answer set altogether and move to other tokens in the vocabulary. The distinction matters because the two changes need not go together, so that a model can improve its relative preference for the correct answer while assigning less probability to that answer in absolute terms, and what the aggregate loss reports as a single loss of competence can in fact contain opposing changes in behaviour.

Which of these changes occurs depends on the displacement that the optimiser applies to the parameters, and with momentum this displacement is not determined by the current gradient alone. Each update combines the current minibatch gradient with a decaying accumulation of earlier gradients, and because these contributions were formed at different parameter states, they need not agree about how the model should move. To first order, the effect of an update on an old task is given by the projection of the applied displacement onto that task's gradient, and nothing in the construction of the update fixes the sign of this projection, so a new-task gradient can locally protect an old behaviour while gradients accumulated earlier in the same training increase its loss.

\begin{figure}[t]
\centering\includegraphics[width=\linewidth]{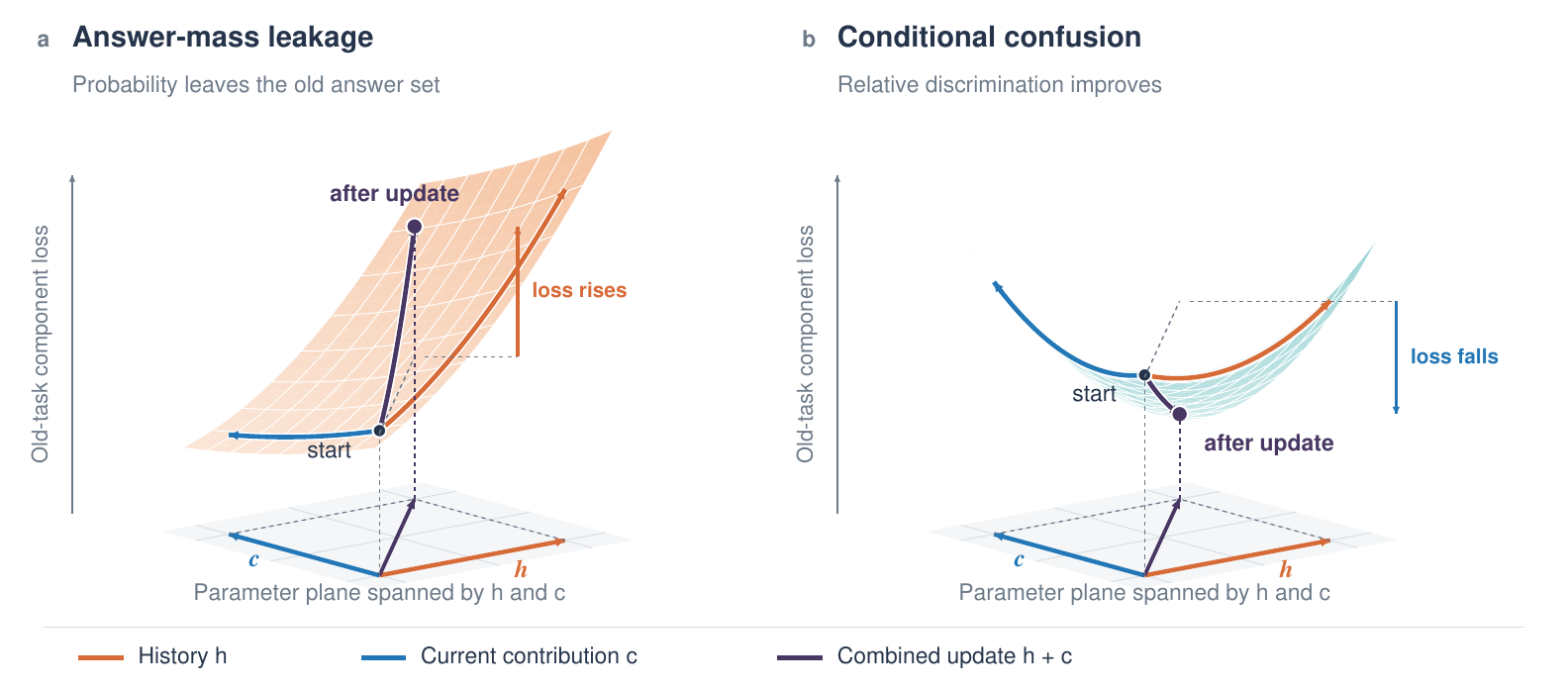}
\caption{\textbf{One update, two changes in retention.} Schematic loss surfaces over the same parameter plane $\theta+u h+v c$, in which the orange and blue curves trace the losses along the history-only and current-only displacements and the purple curve follows their sum. The combined update increases answer-mass leakage while reducing conditional confusion, and curvature bends each trajectory away from its initial slope. Arrows on the floor show the parameter displacements, and vertical arrows compare the loss changes at the endpoints. The surfaces are illustrative, and the reconstruction residual is omitted.}\label{fig:forgettinggeometry}
\end{figure}

We study this interaction by evaluating each contribution to the optimiser update against each of the two components of old-task loss. For an old answer set $S_A$, the factorisation
\[
 p(y\mid x)=p(y\mid x,Y\in S_A)\,p(S_A\mid x)
\]
separates discrimination within the answer set from the total mass assigned to it, and for Adam the first-moment recurrence separates the history carried over from earlier updates from the contribution of the current gradient. Combining the two decompositions gives a component-wise account of each update, which tells us which part of the optimiser moves probability away from the old answers and which part changes their relative ordering, and integrating the corresponding projections along the parameter displacement connects this local description to the finite change in loss that the update actually produces. Figure~\ref{fig:forgettinggeometry} illustrates how a single update can move the two components of the loss in opposite directions.

Across sequential text-classification experiments with three model families, we find that forgetting takes a consistent form, in which the probability assigned to the old answer set falls in every trajectory while discrimination among the old answers usually improves. Within the update, the two optimiser contributions act on this loss in opposite directions, since the accumulated history favours leakage over the course of training while the current gradient opposes it, and when history is resolved by the age of its gradients, the contributions that favour leakage come mainly from older gradients of the new task, whereas the most recent gradients usually act in the opposite direction. The local effect of history on old-task loss changes mainly because the direction of history rotates relative to the old-task gradient, which itself changes much more slowly, rather than because either vector changes in size.

Because these measurements describe associations along natural trajectories, we test them with interventions that change the stored first moment while keeping the starting model, the minibatch, and the size of the first displacement fixed, and we find that changing history alone changes retention, with an immediate effect whose sign depends on the state at which the intervention is made and a longer-term benefit that depends on how far the new task has been learned. Finally, since projections describe an update only locally, we integrate along each finite update and decompose the resulting curvature term into contributions from history, from the current gradient, and from their interaction, which shows that most large increases in old-task loss are already determined by the initial projection, and that in the cases where curvature amplifies the damage, the curvature along history exceeds the protection provided by the mixed history--current term. Taken together, these decompositions and interventions relate changes in retained behaviour to the information that the optimiser accumulates while it learns the new task.

\section{Related work}
Continual-learning methods have long used the sensitivity of earlier tasks to protect acquired behaviour, either by penalising changes along parameter directions that were important for those tasks, as in elastic weight consolidation \citep{kirkpatrick2017overcoming}, or by accumulating the contribution of parameter movement to earlier learning, as in Synaptic Intelligence \citep{zenke2017synaptic}. Gradient Episodic Memory and Orthogonal Gradient Descent instead constrain new updates through the gradients or output sensitivities associated with old tasks \citep{lopezpaz2017gradient,farajtabar2020orthogonal}. Rather than constraining the update, we evaluate the displacement that the optimiser actually produces against separate gradients for confusion and for answer mass, which is what allows us to see how a single update can improve one aspect of retention while degrading another.

Several optimisation methods use stored gradients to shape the update and, in some cases, to preserve earlier behaviour. MoFO selects coordinates according to the magnitude of the momentum \citep{chen2024mofo}, AdEMAMix combines gradient averages at different timescales and finds that older gradients can improve optimisation and slow forgetting \citep{pagliardini2024ademamix}, cautious optimisers use the sign agreement between the update and the current gradient \citep{liang2024cautious}, and \citet{frati2025weight} study how interference during sequential transfer depends on the optimiser. We examine stored gradients through their alignment with a specific component of old-task loss at the state where they are applied, so that the age of a gradient determines its weight in the optimiser's memory, while its functional projection determines whether it favours retention or works against it.

Curvature determines how accurately a local gradient predicts the change in loss at a finite learning rate, an interaction that work on the edge of stability has studied during optimisation \citep{cohen2021edge}, and \citet{li2024revisiting} relate the flatness of the loss landscape to forgetting during the tuning of language models. Our path integrals follow the update that was actually applied and separate its history, current, and mixed curvature contributions, using the same line-integral principle as Integrated Gradients \citep{sundararajan2017axiomatic}, with optimiser channels as the directions of attribution in parameter space.

Competition between old and new outputs is a known source of error in class-incremental learning \citep{wu2019large}, and the vocabulary of a language model makes this competition explicit, since old answers can retain their relative ordering while losing mass to other tokens. The confusion--leakage factorisation connects this output-level behaviour to the geometry of the optimiser, and Appendix~\ref{app:extendedrelated} discusses its links to work on representation change and to the supporting activation interventions.

\section{The geometry of retained behaviour}\label{sec:mechanism}
Let $A$ be an acquired task and $B$ the task currently being learned, write $\theta_t$ for the parameters before update $t$, and write $\delta_t=\theta_{t+1}-\theta_t$ for the applied displacement. Old-task losses and their gradients are always evaluated on a fixed population, so that their changes reflect only the changing state of the model.

\subsection{Two ways to lose an answer}
For a correct answer $y\in S_A$, the full-vocabulary negative log-likelihood satisfies
\begin{equation}
 -\log p_\theta(y\mid x)=
 \underbrace{-\log p_\theta(y\mid x,Y\in S_A)}_{\text{confusion}}
 +\underbrace{-\log p_\theta(S_A\mid x)}_{\text{leakage}},
 \label{eq:answerdecomp}
\end{equation}
and averaging over the population gives $L_A=L_{\rm conf}+L_\ell$, and therefore
\begin{equation}
 g_A=g_{\rm conf}+g_\ell,\qquad
 \Delta L_A=\Delta L_{\rm conf}+\Delta L_\ell,
 \label{eq:componentgradients}
\end{equation}
where $g_F=\nabla L_F$. For soft targets supported on $S_A$, the same factorisation decomposes the full-vocabulary KL divergence into a conditional KL divergence and a leakage term. An increase in $L_\ell$ means that the model assigns less mass to the old answer set, whereas a decrease in $L_{\rm conf}$ means that its predictions within that set improve, and since the total loss reports only their sum, it can hide either change.

The two components have natural counterparts among accuracy measures, since restricted-answer accuracy selects the most probable token within $S_A$ and is therefore invariant to a common rescaling of the answer probabilities, whereas full-vocabulary accuracy selects the most probable token anywhere in the vocabulary and so also depends on how the old answers compete with outside tokens. Equation~\ref{eq:answerdecomp} provides a continuous measurement of both behaviours.

\subsection{Two sources of an update}
For Adam, let $b_t$ be the new-task minibatch gradient after clipping, and let $P_t$ collect the learning rate, the bias correction, and the updated second-moment denominator, so that the first moment and the displacement decompose as
\begin{align}
 m_t&=\beta_1m_{t-1}+(1-\beta_1)b_t,\nonumber\\
 h_t&=-P_t\beta_1m_{t-1},\qquad
 c_t=-P_t(1-\beta_1)b_t,\qquad
 \delta_t=h_t+c_t+r_t.\label{eq:channels}
\end{align}
Here $h_t$ is the history channel, $c_t$ the current channel, and $r_t$ the reconstruction residual of the displacement. Both channels share the same adaptive scaling, so the split isolates the contributions of the first moment, while the memory held in the second moment acts on both of them through $P_t$. Appendix~\ref{app:formalmechanism} gives the explicit denominator and the corresponding decomposition for momentum SGD.

Projecting each optimiser channel onto the gradient of each loss component gives four local contributions,
\begin{equation}
 \mathcal{P}_t=
 \begin{pmatrix}
 g_{{\rm conf},t}^{\top}h_t & g_{{\rm conf},t}^{\top}c_t\\
 g_{\ell,t}^{\top}h_t & g_{\ell,t}^{\top}c_t
 \end{pmatrix},
 \label{eq:pressurematrix}
\end{equation}
in which the rows distinguish confusion from leakage and the columns distinguish history from current learning. A positive entry favours a local increase in loss, the sum of a row gives the first-order change of that behaviour under $h_t+c_t$, and the sum of a column gives the projection of that channel onto the total old-task loss. Opposite signs within a row therefore reveal optimiser contributions that compete over the same behaviour, while opposite signs within a column reveal a single direction that affects the two behaviours differently.

Each entry has the geometric form
\begin{equation}
 g_{F,t}^{\top}u_t=\|g_{F,t}\|\,\|u_t\|\cos\vartheta_{F,u,t},
 \qquad F\in\{{\rm conf},\ell\},\quad u_t\in\{h_t,c_t\},
 \label{eq:localangle}
\end{equation}
so the local projection scales with the norms of both vectors and changes sign with the angle between them. Because $c_t$ is computed from the new task, nothing forces its projection onto an old-task gradient to take a particular sign, and for the same reason a single history vector can increase leakage while decreasing confusion.

Summing over $T$ updates, we define
\begin{equation}
 H_F=\sum_{t=1}^{T}g_{F,t}^{\top}h_t,\qquad
 C_F=\sum_{t=1}^{T}g_{F,t}^{\top}c_t,\qquad
 \Delta L_F=H_F+C_F+E_F,
 \label{eq:cumulative}
\end{equation}
where the cumulative residual $E_F$ collects finite-step effects, the terms $g_{F,t}^{\top}r_t$, and numerical error. We keep this term explicit so that the local contributions can always be compared with the endpoint change that they jointly produce.

\subsection{The age and orientation of history}
Let $m_A$ be the first moment at the task boundary, and number the task-$B$ updates from $t=1$. Expanding the recurrence gives
\begin{equation}
 h_t=\underbrace{-P_t\beta_1^t m_A}_{\text{inherited task-$A$ moment}}
 \; +\sum_{j=1}^{t-1}\underbrace{-P_t(1-\beta_1)\beta_1^{t-j}b_j}_{h_{t\leftarrow j}},
 \label{eq:historyages}
\end{equation}
so that the contribution of gradient $b_j$ to the current leakage sensitivity is
\begin{equation}
 \kappa_\ell(t,j)=g_{\ell,t}^{\top}h_{t\leftarrow j}
 =-(1-\beta_1)\beta_1^{t-j}g_{\ell,t}^{\top}P_tb_j.
 \label{eq:agekernel}
\end{equation}
Since the decay factor is positive, the sign of each contribution is determined by the interaction between the stored gradient, the current preconditioner, and the current leakage sensitivity. We sum these contributions in age bins of 1--4, 5--16, and 17 or more updates, keeping their native magnitudes and bin widths, so that the bins add up to the history channel that was actually applied. Following $\kappa_\ell(t,j)$ at a fixed origin $j$ would, in addition, track how the functional effect of an individual gradient evolves as it ages.

To describe how the aggregate projection changes from one update to the next, we write $a_t=g_t^{\top}u_t=n_td_tz_t$, where $n_t=\|g_t\|$, $d_t=\|u_t\|$, and $z_t$ is the cosine between the two vectors. With $\Delta x_t=x_t-x_{t-1}$ and $\bar x_t=(x_t+x_{t-1})/2$, two exact identities give
\begin{align}
 \Delta a_t&=\underbrace{\bar g_t^{\top}\Delta u_t}_{\text{channel motion}}
 +\underbrace{\Delta g_t^{\top}\bar u_t}_{\text{task-gradient motion}},\label{eq:symmetricmotion}\\
 \Delta a_t&=\underbrace{\overline{nd}_t\,\Delta z_t}_{\text{orientation}}
 +\underbrace{\bar z_t\bar d_t\,\Delta n_t}_{\text{gradient norm}}
 +\underbrace{\bar z_t\bar n_t\,\Delta d_t}_{\text{channel norm}},\label{eq:symmetricpolar}
\end{align}
where $\overline{nd}_t=(n_td_t+n_{t-1}d_{t-1})/2$. The first identity separates the motion of the channel from the motion of the task gradient, and the second separates changes in orientation from changes in either norm. Within each identity, we measure the share of each term in the summed absolute scalar contributions, which avoids cancellation across updates while keeping the vector-motion and magnitude--orientation attributions separate. Appendix~\ref{app:historydynamics} defines the normalisation, together with a nested decomposition of channel motion into changes in the numerator and in the adaptive scaling.

\section{Experimental setting}\label{sec:setting}
We study SST-2 sentiment classification followed by AG News topic classification \citep{socher2013sentiment,zhang2015text}, with each task posed as a plain-text instruction that the model answers with a single token. Each task has 1,024 training, 32 validation, and 128 test texts, with prefixes truncated to 256 tokens. Pythia-410M, SmolLM2-135M, and Qwen3-0.6B-Base each contribute three seeds under two answer-codebook conditions, which gives 18 trajectories in total. Within a seed, the shared and disjoint codebooks start from the same acquired old-task state and use the same underlying data, and they differ only in the labels of the new task, which change together with its instructions.

Each model learns task $A$ for 600 updates and then task $B$ for 128 updates, with the Adam state carried across the boundary, and all 18 trajectories meet the acquisition criterion for the old task. Adam uses $\beta_1=.9$, $\beta_2=.99$, $\epsilon=10^{-8}$, global clipping at norm one, and no weight decay. The learning rates are $5\times10^{-6}$ for Pythia and $8\times10^{-5}$ for SmolLM2, selected by the validation calibration described in Appendix~\ref{app:optimizer}, while Qwen uses $2\times10^{-5}$, taken from the association-learning protocol. Development and confirmation pools are separated by hashing the texts, and seeds may share examples within a pool.

Behavioural endpoints are measured on all 128 test texts, whereas the history replay, which recomputes component gradients and outcomes at every update, uses a fixed subset of 32 test texts, chosen by their position in the test order and therefore independently of harm, together with all 32 validation texts, and it covers seven Pythia and SmolLM2 trajectories and all six Qwen trajectories, for a total of 1,664 updates. The seven Pythia and SmolLM2 cases are the first to have completed, so they are not balanced across conditions, and their reconstructed parameter states and channel sums pass the checks described in Appendices~\ref{app:historydynamics} and~\ref{app:newreplication}.

The history interventions are run as separate optimiser studies in which the compared branches share their starting weights, their minibatches, and all optimiser state other than history. Immediate probes compare preserved history with a reset at fixed steps, while continuation experiments branch at updates 8, 32, and 64 and train for a further 32 steps, so that both retention and new-task learning can be compared, and in both cases global or blockwise norm matching constrains the first modified displacement. Appendices~\ref{app:optimizer} and~\ref{app:followup} specify the populations and the acquisition criteria.

The finite-path measurements use adaptive integration together with independent checks of the derivatives. Pythia and SmolLM2 contribute 252 validation and test population paths, of which 204 pass every component check, and Qwen contributes 24 paths, all of which pass. When we report how often an event occurs, we keep uniformly sampled updates separate from events selected on validation loss, and Appendix~\ref{app:newpaths} gives the numerical criteria, the unresolved cases, and the selection rules. The supporting experiments on association learning, forecasting, and activation interventions are described in Appendices~\ref{app:movedresults}--\ref{app:optimizer}.

\section{History and current learning act on answer mass in opposite directions}\label{sec:results}
\subsection{Answer mass falls while conditional predictions improve}
\begin{figure}[t]
\centering\includegraphics[width=\linewidth]{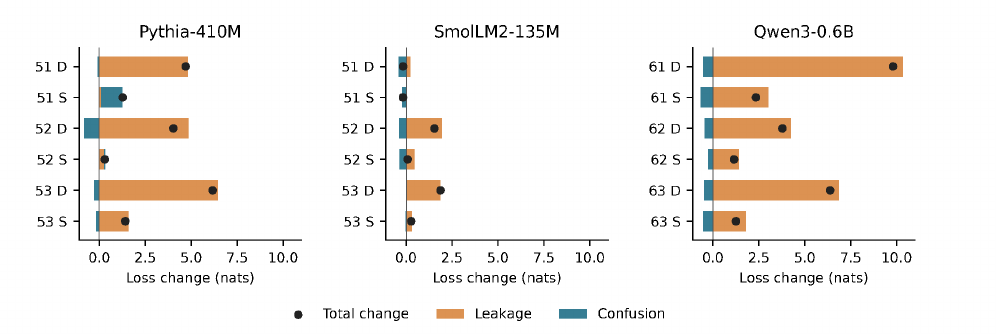}
\caption{Two forms of retention change after 128 new-task updates. Leakage increases in all 18 trajectories, while conditional confusion decreases in 16 of them. Bars show the two loss components on the full 128-text test population and black markers show their sum, with D/S denoting disjoint and shared answer codebooks, which are paired within seed.}\label{fig:functionaloverview}
\end{figure}
Figure~\ref{fig:functionaloverview} shows that leakage increases in every trajectory, while conditional confusion decreases in 16 of the 18, and in two SmolLM2 runs the gain in conditional prediction is large enough to outweigh the loss of answer mass, so that the total old-task loss actually improves. A single aggregate loss would therefore conceal the one behavioural change that all runs have in common.

The same separation appears in accuracy, where Pythia retains substantial restricted-answer accuracy even when its full-vocabulary accuracy falls to zero. Under disjoint codebooks, all three Pythia seeds end with zero full-vocabulary accuracy while their restricted-answer accuracy remains at 63.28\%, 78.12\%, and 81.25\% (Table~\ref{tab:finalaccuracy}), which means that the correct answer often still outranks the other old answer while some outside token outranks both. Full-vocabulary accuracy falls in all twelve Pythia and SmolLM2 runs, including the two whose total loss improves.

In all nine paired comparisons of the natural-text study, disjoint codebooks produce greater leakage and greater total deterioration than shared ones. Since token identities and instructions change together between the two conditions, this comparison measures their combined effect on transfer, and the seeds of the optimiser study show more variation, including one pair in which the ordering of leakage is reversed (Table~\ref{tab:label_control}). What all these comparisons have in common is the separation between discrimination within the answer set and the probability of selecting an old answer at all.

\subsection{A consistent sign pattern within the update}
\begin{table}[htbp]
\centering\small
\caption{History and current-gradient contributions on the fixed 32-text $A$-test subset. Values are sums over 128 updates in nats. $H_\ell=\sum_t g_{\ell,t}^{\top}h_t$ and $C_\ell=\sum_t g_{\ell,t}^{\top}c_t$ sum the local leakage projections along the recorded trajectory. $E_\ell=\Delta L_\ell-H_\ell-C_\ell$ retains finite-step and displacement/numerical residuals.}\label{tab:historyleakage}
\setlength{\tabcolsep}{3pt}
\begin{tabular}{llrrrrr}\toprule
Model & Seed / answers & $\Delta L_\ell$ & $H_\ell$ & $C_\ell$ & $E_\ell$ & $\Delta L_{\rm conf}$\\\midrule
Pythia-410M & 51 / disjoint & +4.540 & +4.591 & -0.675 & +0.624 & -0.020\\
Pythia-410M & 51 / shared & +0.084 & +0.176 & -0.225 & +0.134 & +1.339\\
Pythia-410M & 52 / shared & +0.136 & +0.667 & -1.131 & +0.599 & +0.293\\
SmolLM2-135M & 51 / disjoint & +0.217 & +0.277 & -0.150 & +0.089 & -0.650\\
SmolLM2-135M & 51 / shared & +0.045 & +0.078 & -0.072 & +0.039 & -0.505\\
SmolLM2-135M & 52 / disjoint & +1.878 & +1.859 & -0.195 & +0.214 & -0.387\\
SmolLM2-135M & 52 / shared & +0.241 & +0.305 & -0.145 & +0.082 & -0.363\\
Qwen3-0.6B & 61 / disjoint & +10.536 & +11.183 & -2.709 & +2.062 & +0.351\\
Qwen3-0.6B & 61 / shared & +3.006 & +3.332 & -1.238 & +0.912 & -0.373\\
Qwen3-0.6B & 62 / disjoint & +4.369 & +5.269 & -2.605 & +1.705 & +0.103\\
Qwen3-0.6B & 62 / shared & +1.250 & +2.326 & -2.198 & +1.123 & +0.155\\
Qwen3-0.6B & 63 / disjoint & +6.986 & +7.016 & -0.944 & +0.914 & -0.356\\
Qwen3-0.6B & 63 / shared & +1.646 & +2.250 & -1.185 & +0.581 & -0.419\\
\bottomrule\end{tabular}
\end{table}

In all 13 replays, history and the current gradient have cumulative leakage projections of opposite sign,
\begin{equation}
 H_\ell>0,\qquad C_\ell<0,\qquad \Delta L_\ell>0,
 \label{eq:empiricalsigns}
\end{equation}
so that, summed over training, the history channel points locally towards a reduction of old-answer mass, while the current channel points towards its recovery (Table~\ref{tab:historyleakage}). Both channels originate from the same training process, and their different signs appear only when their displacements are evaluated against the leakage gradient of the old task.

Although this sign pattern is shared across replays, the cumulative local sum of the two channels can either favour leakage or oppose it. In Qwen seed 61 with disjoint answers, history contributes $+11.183$ nats and the current channel $-2.709$ nats to a leakage increase of $10.536$ nats, whereas in Pythia seed 52 with shared answers the local sum is negative, $0.667-1.131=-0.464$ nats, even though leakage increases by $0.136$ nats, because a cumulative residual of approximately $+0.599$ nats reverses the sign of the endpoint account. Equation~\ref{eq:cumulative} keeps this accumulated departure from the local linearisation visible.

The relationship between history and confusion, in contrast, differs across models, since the cumulative confusion projections of history are harmful in the three Pythia replays and beneficial in the four SmolLM2 replays, so that one optimiser channel can favour leakage and improved discrimination at the same time. The replay values are computed on matched 32-text populations, while the endpoint values in Figure~\ref{fig:functionaloverview} average over the full 128-text test set, and this difference matters for confusion, which rises in three of the Qwen replay subsets even though it falls at all six full-test endpoints, showing how sensitive this component is to the evaluation population.

\subsection{The positive history contributions come from older new-task gradients}
\begin{figure}[t]
\centering\includegraphics[width=\linewidth]{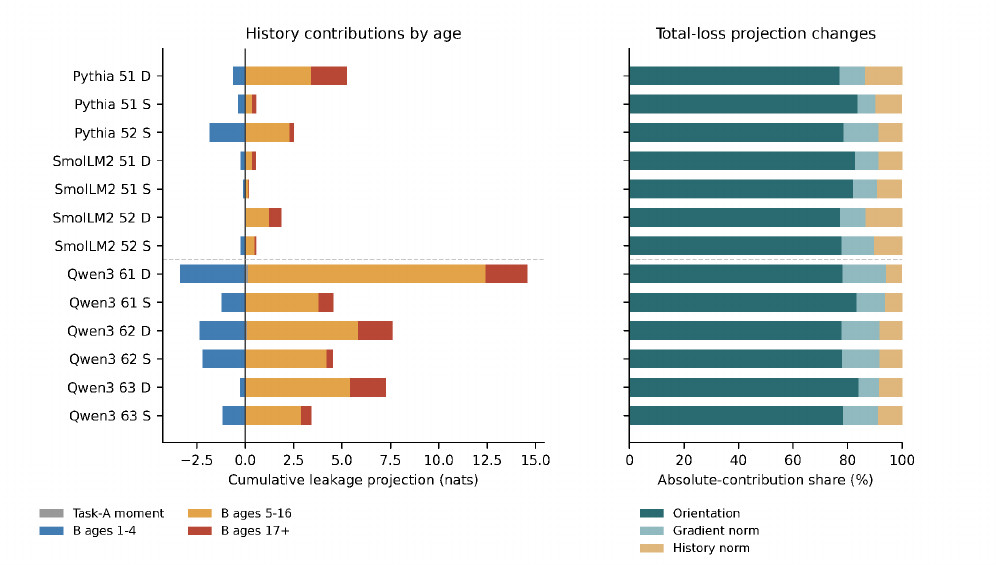}
\caption{Age and orientation of optimiser history. The left panel shows cumulative leakage projections by origin and age, where positive values favour leakage, and the right panel shows the absolute-contribution shares for changes in the total-loss history projection, computed with Equation~\ref{eq:symmetricpolar}. The left panel therefore describes accumulated leakage contributions, while the right panel describes how the total-loss projection varies across updates.}\label{fig:historydynamics}
\end{figure}
Expanding $h_t$ according to Equation~\ref{eq:historyages} shows where its positive leakage contributions come from. Both of the older task-$B$ bins, with ages 5--16 and 17 or more, have positive cumulative projections in every replay, whereas ages 1--4 contribute negatively in twelve of the thirteen, and the moment inherited from task $A$ makes a comparatively small net contribution (Figure~\ref{fig:historydynamics}; Table~\ref{tab:historyages}). The opposition between history and the current gradient therefore extends into history itself, with the most recent gradients sharing the sign of the current gradient and the older gradients carrying the opposite sign. In Pythia seed 51 with disjoint answers, for example, ages 1--4 contribute $-0.649$ nats, ages 5--16 contribute $+3.382$, and ages 17 and above contribute $+1.841$, while the terminal task-$A$ moment contributes only $+0.017$, so that almost all of the positive net projection comes from history accumulated while learning the new task, and a reset confined to the task boundary would leave these later contributions intact.

These sums measure the contributions that are actually present in the optimiser, including their decay, their magnitude, and the width of each bin. Equation~\ref{eq:agekernel} clarifies how to read them, since age attenuates a stored vector while its projection determines the sign of its effect, and the pattern across bins therefore motivates following individual gradients as their functional alignment changes. Qwen's records of fixed-origin cohorts provide a basis for such an analysis (Appendix~\ref{app:newreplication}), whereas the result reported here concerns the cumulative contributions of the age groups.

\subsection{Orientation dominates changes in history's local effect}
\begin{table}[htbp]
\centering\small
\caption{Changes in history's total-loss projection. Ranges across trajectories, in percent. Orientation uses the three-term decomposition in Equation~\ref{eq:symmetricpolar}; history motion uses Equation~\ref{eq:symmetricmotion}; adaptive scaling is nested within history motion. The columns have separate denominators.}\label{tab:alignmentsummary}
\setlength{\tabcolsep}{3pt}
\begin{tabular}{lrrrr}\toprule
Model & $n$ & Orientation & History motion & Adaptive scaling\\\midrule
Pythia-410M & 3 & 77.19--83.69 & 65.71--75.56 & 1.82--3.21\\
SmolLM2-135M & 4 & 77.35--82.77 & 68.03--73.41 & 1.80--2.62\\
Qwen3-0.6B & 6 & 77.89--83.97 & 59.10--73.56 & 1.84--3.23\\
\bottomrule\end{tabular}
\end{table}

For the total old-task loss, the orientation term in Equation~\ref{eq:symmetricpolar} accounts for 77.19--83.97\% of the summed absolute terms across replays, and it is the largest of the three terms in every trajectory (Table~\ref{tab:alignmentsummary}), so the functional effect of history varies mainly because its direction changes relative to the task gradient.

The vector decomposition in Equation~\ref{eq:symmetricmotion} attributes more of this change to the motion of history than to the motion of the task gradient, with channel motion accounting for 59.10--75.56\% of the absolute two-term attribution. In the Pythia and SmolLM2 replays, the median cosine between consecutive history vectors is approximately $0.90$, while the median cosine between consecutive task gradients is $0.998$--$0.999$ (Table~\ref{tab:historymotion}), so the aggregate direction of history evolves against a task gradient that rotates much more slowly.

Changes in the adaptive denominator contribute only 1.80--3.23\% within the nested decomposition of history motion, a share that measures variation around the native Adam preconditioner, which shapes both channels at every step. Since each of the three reported shares uses the normalisation of its own identity, they are not additive, but taken together they place the observed changes in the projection mainly in the orientation of the evolving first-moment contribution.

\section{Changing history changes retention}\label{sec:interventions}
The projection analysis suggests a direct intervention, in which the stored first moment is changed while the starting model and the minibatch are kept fixed. Resetting history changes the direction of the proposed update, and matching its norm to that of the original update then tests the effect of this change in direction at the original scale of movement.

Let $q_t$ be the displacement proposed after resetting the first moment, with the second moments and the optimiser clock preserved. For each parameter block $b$ with a nonzero reset displacement, we define
\begin{equation}
 \widetilde\delta_t^{(b)}=
 \frac{\|\delta_t^{(b)}\|}{\|q_t^{(b)}\|}\,q_t^{(b)},
 \qquad
 \|\widetilde\delta_t^{(b)}\|=\|\delta_t^{(b)}\|,
 \label{eq:matchedreset}
\end{equation}
so that a single block gives global matching, while several blocks preserve the distribution of displacement norms across the specified parameter groups. The paired loss difference
\begin{equation}
 D_F=L_F(\theta_t+\widetilde\delta_t)-L_F(\theta_t+\delta_t)
 \label{eq:resetoutcome}
\end{equation}
then measures the immediate effect of this change in the direction of the update, with $D_F<0$ favouring the reset, and the experiments verify the matching on the displacement that is actually assigned, including differences due to finite precision.

\begin{table}[htbp]
\centering\small
\caption{Illustrative fixed-step SmolLM2 history probes. Entries are immediate changes in test loss, and the reset displacement norms match the preserved norms within each block.}\label{tab:history_control}
\setlength{\tabcolsep}{3pt}
\begin{tabular}{lrrrrrr}\toprule
Optimiser & Seed & Step & $A$: keep & $A$: reset & $B$: keep & $B$: reset\\\midrule
Momentum SGD & 23 & 8 & +0.100279 & -0.275243 & -0.218030 & -0.411795\\
Adam & 21 & 8 & +0.079993 & -0.035992 & +0.040147 & -0.911440\\
Adam & 22 & 8 & -0.038177 & -0.012278 & -0.271431 & -0.049603\\
\bottomrule\end{tabular}
\par\smallskip\footnotesize These probes from the optimiser study evaluate the immediate modified update, starting from the same initial weights, minibatch, optimiser clock, and, where applicable, second moments, and the displayed cases illustrate how the direction of the intervention effect varies.
\end{table}

The immediate effect depends on the state at which the intervention is made (Table~\ref{tab:history_control}). At update 8 of SmolLM2 Adam seed 21, preserving history increases old-task loss by $0.080$ nats while a block-matched reset decreases it by $0.036$, whereas at update 8 of seed 22 preserving history gives the larger improvement, and momentum SGD also shows a substantial effect of the matched reset. Under both optimisers, the information stored in the first moment can therefore either help or harm retention at a fixed scale of displacement.

\begin{table}[htbp]
\centering\small
\caption{Differences between first-step block-matched history-reset branches and preserved-history branches, given as mean final test-loss differences after 32 updates, where the matching constrains only the first displacement. The numbers of acquired trajectories and of forks are shown separately.}\label{tab:followupbranches}
\setlength{\tabcolsep}{3pt}
\begin{tabular}{llrrrrrr}\toprule
Model & Optimiser & Seeds & Forks & $\Delta A_f$ & $\Delta B_f$ & Confusion & Leakage\\\midrule
Pythia-410M & Adam & 3 & 9 & $-.127170$ & $-.001890$ & $+.099843$ & $-.227013$\\
SmolLM2-135M & Adam & 3 & 9 & $-.266907$ & $+.001482$ & $-.071779$ & $-.195129$\\
SmolLM2-135M & Momentum & 2 & 6 & $-.085724$ & $+.030098$ & $-.069111$ & $-.016613$\\
\bottomrule\end{tabular}
\end{table}

To measure retention after further learning of the new task, we branch at updates 8, 32, and 64 and continue each branch for 32 steps. Averaging the forks within each acquired trajectory, block-matched resets give a lower final old-task loss in all six Adam model and seed trajectories, and Table~\ref{tab:followupbranches} reports these differences together with the new-task loss and the two components of retention.

For Pythia, leakage improves by $0.227$ nats while confusion worsens by $0.100$, which gives a net improvement of $0.127$ nats in old-task loss, and full-vocabulary accuracy rises by 6.337 percentage points while restricted-answer accuracy changes by only 0.087 points, so the intervention mainly restores the probability of producing an old answer. In SmolLM2, both components of the loss improve on average, which shows that the same intervention can recover answer mass while producing different changes in conditional discrimination.

Norm matching constrains only the first displacement, after which the parameters and optimiser states of each branch evolve independently, so the benefit of a reset also depends on how far each branch has progressed on the new task. When the branches are compared at equal new-task validation loss, the mean Pythia difference in old-task loss is $+0.003580$ nats at a target of $0.75$ and $-0.025831$ nats at a target of $1.0$, and matching on acquisition in this way complements the fixed-step comparison when judging the benefit of a history reset.

\section{From local projections to finite loss changes}\label{sec:finite}
A local projection evaluates sensitivity at $\theta_t$, but the finite update traverses the segment $\theta_t(\alpha)=\theta_t+\alpha\delta_t$, along which the sensitivity can change. For any task or component loss $L_T$, the fundamental theorem of calculus gives
\begin{equation}
 \Delta L_T=\int_0^1 s_T(\alpha)\,\dd\alpha
 =g_T^\top\delta_t+R_T,\qquad
 s_T(\alpha)=\nabla L_T(\theta_t(\alpha))^\top\delta_t,
 \label{eq:path}
\end{equation}
and the same path yields a finite attribution to each channel,
\begin{equation}
 I_T^u=\int_0^1\nabla L_T(\theta_t(\alpha))^\top u_t\,\dd\alpha,
 \qquad \Delta L_T=I_T^h+I_T^c+I_T^r,
 \label{eq:integratedchannelsmain}
\end{equation}
in which the channel vectors stay fixed while the gradient evolves along the segment, so that their integrals divide the loss change among the channels over the full displacement. A history intervention, by contrast, changes the displacement itself, and with it the path along which Equation~\ref{eq:path} is evaluated.

\subsection{Curvature resolves the local approximation error}
For twice-differentiable $L_T$, we define the weighted path Hessian
\begin{equation}
 \mathcal K_T=\int_0^1(1-\alpha)H_T(\theta_t(\alpha))\,\dd\alpha,
 \qquad R_T=\delta_t^\top\mathcal K_T\delta_t,
 \label{eq:curvature}
\end{equation}
where $H_T=\nabla^2L_T$, and substituting $\delta_t=h_t+c_t+r_t$ gives
\begin{align}
 R_T={}&\underbrace{h_t^\top\mathcal K_T h_t}_{R_T^{hh}}
 +\underbrace{c_t^\top\mathcal K_T c_t}_{R_T^{cc}}
 +\underbrace{h_t^\top\mathcal K_T c_t+c_t^\top\mathcal K_T h_t}_{M_T}
 +R_T^{\rm res},\label{eq:curvaturesplit}
\end{align}
where the residual $R_T^{\rm res}$ collects the terms involving $r_t$. The mixed term $M_T$ can either reinforce or oppose the diagonal contributions, and in particular $R_T>0$ and $M_T<0$ can hold together, in which case the full step amplifies the loss while the interaction between the channels partially protects it. Since $\mathcal K_T$ enters only through directional contractions, the computation never requires the full Hessian explicitly.

\subsection{Most sampled large increases are first-order dominated}
\begin{table}[htbp]
\centering\small
\caption{Uniform test paths on fixed 32-text subsets. P/S: Pythia/SmolLM2, 96 paths per component; Q: Qwen3, six. Acceptance is component-specific. Large increases exceed 0.05 nats; FO and Amp. follow the ordered rule. Mix counts material mixed terms across all accepted paths. Sampling densities differ between studies.}\label{tab:coverage}
\setlength{\tabcolsep}{3pt}
\begin{tabular}{llrrrrrr}\toprule
Study & Component & Accepted & Unresolved & Large & FO & Amp. & Mix\\\midrule
P/S & Total & 87 & 9 & 19 & 17 & 2 & 1\\
P/S & Confusion & 81 & 15 & 5 & 5 & 0 & 1\\
P/S & Leakage & 89 & 7 & 15 & 13 & 2 & 0\\
Q & Total & 6 & 0 & 2 & 2 & 0 & 0\\
Q & Confusion & 6 & 0 & 0 & 0 & 0 & 0\\
Q & Leakage & 6 & 0 & 2 & 2 & 0 & 0\\
\bottomrule\end{tabular}
\end{table}

We classify increases of more than $0.05$ nats using the initial projection $P=g_T^\top\delta_t$ and the remainder $R=\Delta-P$. An event counts as a favourable-to-harmful reversal when $P<-0.01$, and among the remaining large events, amplification requires $R>0.01$ and $R\geq0.25|\Delta|$, while first-order dominance then requires $P>0$ and $|P|\geq|R|$. Separately, the mixed curvature is considered material when $|M|\geq0.01$ and $|M|\geq0.25|\Delta|$. These operational thresholds separate the size of the remainder from the sign and size of its mixed part.

Of the 19 large total-loss increases that pass the checks in the uniform Pythia and SmolLM2 test sample, 17 are first-order dominated and two show amplification, and in Qwen, both large increases in the uniform sample and all five in the validation-selected test sample are first-order dominated. No accepted path on the natural components reverses a favourable initial direction. Table~\ref{tab:coverage} also lists the unresolved paths, and validation and test evaluations at the same update, as well as paired codebook conditions, are not independent of each other.

\subsection{History curvature can outweigh protective coupling}
The two Pythia amplification paths occur at update 9 of seed 53 under the paired codebooks, with total-loss increases of $0.099559$ and $0.139375$ nats and remainders of $0.050205$ and $0.041175$ nats, and Table~\ref{tab:curvatureexamples} resolves their directional curvature contributions.

\begin{table}[htbp]\centering\small
\caption{Weighted curvature contributions to total old-task loss at update 9 of Pythia seed 53. History curvature is positive and mixed curvature is protective under both codebooks. Values are in nats, and the closure for the full displacement is given in Table~\ref{tab:finalcurvature}.}\label{tab:curvatureexamples}
\begin{tabular}{lrrr}\toprule
Answers & $R_A^{hh}$ & $R_A^{cc}$ & $M_A$\\\midrule
Disjoint & $+0.055557$ & $+0.000814$ & $-0.006253$\\
Shared & $+0.082203$ & $+0.013482$ & $-0.054475$\\\bottomrule
\end{tabular}
\end{table}

In both paths, positive curvature along history is the largest contribution, and although mixed curvature offsets it, strongly so under shared answers, the diagonal terms remain larger in combination. The independently integrated curvature of the full displacement agrees with each remainder to about $2.8\times10^{-5}$ nats (Table~\ref{tab:finalcurvature}), so the positive remainder in these paths reflects curvature along history that exceeds the protection provided by the mixed term.

The Qwen paths also contain mixed curvature of both signs, although none of them meets the materiality criterion. At update 73 of seed 61 with shared answers, $M_A<0$ while $R_A>0$, whereas at update 12 of seed 63 with disjoint answers, $M_A>0$ and contributes about 7.7\% of the loss increase (Table~\ref{tab:qwenpathtotal}), so the signed decomposition remains informative even below the classification thresholds.

The supporting association-learning study contains an update whose old-task loss rises despite a favourable initial projection. At the highlighted OLMo step, $P=-0.130264$ nats but $\Delta L_A=+0.287872$, with integrated history contributing $+0.309796$ and the current channel $-0.021920$ (Figure~\ref{fig:paths}), so in this case the finite-step remainder reverses the sign of the initial projection. In the larger natural-text sample, however, the accepted large increases are predominantly first-order dominated, and Appendix~\ref{app:movedpath} follows this reversal in detail.

\section{Discussion and Limitations}
Optimiser history affects both whether an old answer is produced and how it is ranked among the old alternatives, and this distinction matters when assessing recovery, since the Pythia reset branches recover answer mass despite worse conditional discrimination, while two natural SmolLM2 runs improve their total loss despite losing answer mass. Evaluating both components exposes these trade-offs and connects the effect of an optimiser intervention to the behaviour that it actually preserves.

Our empirical findings concern sequential classification with single-token answers, a fixed task order, and model-specific learning rates. The recurrence of the results across the tested model families, seeds, and codebooks supports their consistency within this setting, but their extension to open-ended generation remains untested. The codebook conditions are paired and seeds can share evaluation examples, so these runs are not fully independent, and while the 32-text replay populations allow matched gradient attribution, the 128-text endpoints characterise broader behavioural change, which is why the trends in conditional confusion sometimes differ between the two.

The age decomposition measures the stored contributions that are active at each state, including their decay, magnitude, and bin width, so it establishes which age groups supply the positive leakage projections without isolating staleness as their cause. The orientation shares depend on the native parameterisation and on the stated normalisation, and the matched-history interventions establish an effect of stored first-moment information at a controlled initial scale of displacement, while their longer-term benefit for retention depends on how the new task is subsequently acquired.

Our conclusions about finite paths rely on the trajectories that pass the numerical checks, and the 48 unresolved Pythia and SmolLM2 population paths, together with the single uniformly sampled Qwen update per trajectory, limit our estimates of how often each kind of event occurs. The accepted paths nevertheless resolve how history, current learning, and mixed curvature combine in individual loss changes, so they support the measured interactions between channels, while their prevalence in the population remains constrained by the sampling and acceptance criteria.

\section{Conclusion}
Forgetting can reduce the probability that a model produces an old answer while preserving, or even improving, its discrimination within the old answer set, and in our sequential-task experiments this change is accompanied by a consistent opposition within Adam's update, in which the cumulative projections of history favour leakage while those of the current gradient oppose it. Resolving history by age places the positive contributions in older gradients of the new task, and interventions at matched states connect the information stored in the optimiser to the recovery of answer mass. Integrating along finite paths then measures how sensitivity changes along each displacement and identifies cases in which curvature amplifies the increase in loss. Together, these measurements relate the loss of answer mass to stored new-task gradients, to their changing orientation, and to their interaction with the current update.
\label{main-end}

\clearpage
\appendix
\section{Supporting registry and activation analyses}\label{app:movedresults}
The association-learning study follows individual optimiser updates and tests whether activation interventions recover behaviour at their endpoints, complementing the text-classification analysis of history with controlled soft targets and held-out prompt templates.

\subsection{Activation orientation at fixed channel states}
\label{app:movedframe}
\label{sec:frame}
For OLMo, we test four selected updates at the zero-indexed layers 0, 1, and 9. At each layer, we fit a proper orthogonal Procrustes map from the post-update to the pre-update final-token residual activations, using validation prompts from 16 calibration entities across both tasks, and the fit is uncentred and restricted to the joint row span of the calibration activations, with the identity acting outside that span. Test prompts use the other 16 entity identities, and the map, which is fitted only from paired calibration activations, is held fixed across tasks, update fractions, and channel conditions.

For $u,v\in\{0,1\}$, we evaluate the four parameter states
\begin{equation}
 \theta_{uv}(\alpha)=\theta_t+\alpha(uh_t+vc_t+uvr_t),
 \qquad \alpha\in\{0.25,0.625,1\},
 \label{eq:corners}
\end{equation}
where the state $11$ agrees with the natural interpolation up to recorded rounding. At each state, we patch only the final-token residual output of the selected layer and then run the rest of the model. The controls include a sham patch, the inverse rotation, a mean shift and a scalar scaling derived from the calibration set, four rotations with matched eigenangles in random subspaces, and four additive perturbations whose norm matches, for each prompt, the displacement produced by the fitted rotation at that state.

We measure endpoint rescue by comparing patched and unpatched loss and accuracy, and we quantify the interaction between the channels through $J_T=L_T(\theta_{11})-L_T(\theta_{10})-L_T(\theta_{01})+L_T(\theta_{00})$, which measures their joint effect relative to their separate effects. All four states are evaluated with the recorded fixed preconditioner and displacement channels. Geometry is measured on the 16 test activations, for which we compute the centred covariance spectrum and the $H_0/H_1$ Vietoris--Rips persistence from Euclidean distances normalised by the median pre-update distance, using simplices up to dimension two over $\mathrm{GF}(2)$.

\subsection{A reversal along the update path}
\label{app:movedpath}
\begin{figure}[t]
\centering
\includegraphics[width=\linewidth]{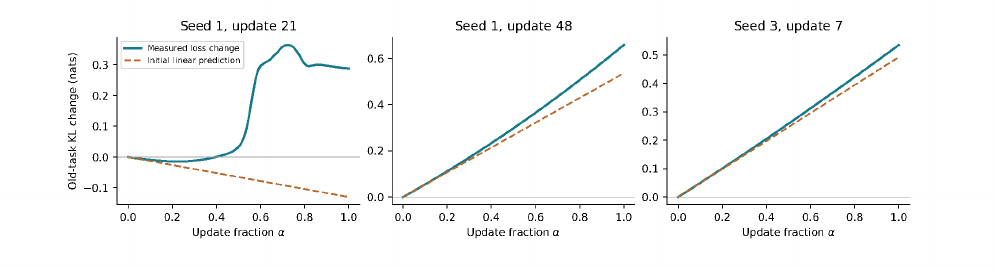}
\caption{Old-task loss along diagnostic interpolations of three recorded OLMo displacements, with dashed lines showing the initial linear prediction. The event on the left reverses its initially favourable direction, while the other events largely follow their initial predictions. All displayed paths pass their quadrature checks, and update 21 uses the refined evaluations described in Appendix~\ref{app:audits}.}
\label{fig:paths}
\end{figure}
The highlighted OLMo reversal shows the sensitivity of the task changing within a single step (Figure~\ref{fig:paths}). Both the history and the current-gradient channels initially point towards lower old-task loss, but farther along the displacement the loss rises sharply, and the accumulated nonlinear contribution outweighs this favourable beginning.

Integration shows where this damage accumulates, with history making the dominant positive contribution to the change in old-task loss while the current-gradient channel remains beneficial overall (Table~\ref{tab:channels_main}). Since the channels themselves stay fixed along the interpolation, their changing effect comes entirely from the evolving task gradient, which turns an initially helpful history projection into a harmful integrated contribution.

The decomposition of the remainder describes this evolution in more detail, since variation in the gradient magnitude offsets part of the damage, while variation in orientation and its interaction with magnitude both contribute positively, and the large joint term shows how the two changes reinforce each other (Table~\ref{tab:writing_path_details}). Quadrature, slope, and sampled curvature checks support the reported decomposition of this path, and Appendix~\ref{app:audits} gives the full numerical audit.

\subsection{Channel contributions across events}
\label{app:movedchannels}

\begin{table}[t]\centering\small
\caption{Selected audited old-task paths, all from seed 1. The integrated channel contributions reconstruct the total change, and the small residual and quadrature errors are omitted from this table.}\label{tab:channels_main}
\setlength{\tabcolsep}{4pt}
\begin{tabular}{lrrrrr}\toprule
Checkpoint & Update & $\Delta L_A$ & $s_A(0)$ & $I_A^h$ & $I_A^c$\\\midrule
OLMo-1B & 21 & +0.287872 & -0.130264 & +0.309796 & -0.021920\\
Pythia-70M & 61 & +0.124563 & +0.102805 & +0.102987 & +0.021562\\
Pythia-410M & 65 & +0.322116 & +0.248560 & +0.231985 & +0.090131\\
Qwen2.5 Instruct & 107 & +0.161973 & +0.155524 & +0.042224 & +0.119749\\
SmolLM2 base & 73 & +0.095240 & +0.082083 & +0.047804 & +0.047435\\
SmolLM2 Instruct & 64 & +0.106675 & +0.101283 & +0.072903 & +0.033771\\
Qwen3 base & 3 & +0.000294 & -0.000029 & -0.000307 & +0.000601\\
Qwen3 base & 59 & +0.075012 & +0.065861 & +0.061498 & +0.013514\\
\bottomrule\end{tabular}\end{table}

The other audited paths reach harmful endpoints through different balances between the channels, since history dominates the selected Pythia-70M event, the current-gradient channel dominates the selected Qwen2.5 Instruct event, and the two contribute comparably in the selected SmolLM2 base event (Table~\ref{tab:channels_main}). In the other OLMo paths of Figure~\ref{fig:paths}, the initial projection already explains most of the deterioration, and variation in the gradient magnitude supplies the main positive contribution to the smaller remainder.

Qwen3 shows the same variation within a single model, since in its larger selected event history drives the damage to the old task and outweighs an improving current-gradient contribution on the new task, so that both held-out losses rise, while the changing orientation offsets part of the old-task damage, and in its smaller event history protects retention while the current channel increases old-task loss (Table~\ref{tab:writing_qwen_details}). In each case, the role of a contribution follows its alignment with the task at the state being traversed.

The Pythia text study also contains a resolved reversal, in which a favourable initial slope on the old task is outweighed by a positive finite-step remainder, and its selected paths include both harmful and protective integrated history, with their integration and sampled curvature checks reported in Appendix~\ref{app:extensions} and Table~\ref{tab:text_path_details}. Across these suites, the decomposition of Section~\ref{sec:mechanism} provides a common way to explain the differing outcomes.

\subsection{Activation rescue depends strongly on the model state}
\label{app:movedrescue}
\begin{figure}[t]
\centering
\includegraphics[width=\linewidth]{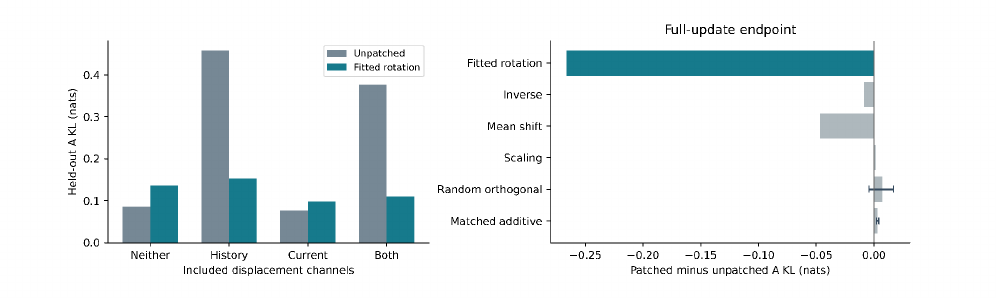}
\caption{The highlighted OLMo reversal, evaluated at layer 0 on entities held out from the fitting of the rotation. The left panel shows old-task losses at the four fixed channel states before and after the fitted rotation, and the right panel shows endpoint loss changes under the fitted map and the comparison conditions, where the bars for random controls give means and ranges across four realisations. Negative changes indicate improved performance.}
\label{fig:frame}
\end{figure}
At the full-update endpoint of the highlighted OLMo reversal, a fitted activation rotation recovers most of the increase in old-task loss and restores two-answer accuracy to its pre-update level, with most held-out entities improving and the new task benefiting as well (Figure~\ref{fig:frame}; Table~\ref{tab:writing_rescue_details}), so the recovery extends to entities that were excluded when the map was fitted.

At this endpoint, the fitted map is more effective than its inverse, than the mean shift and scaling derived from the calibration set, and than every sampled random rotation and matched additive perturbation (Table~\ref{tab:writing_rescue_details}), and this advantage over the controls links the observed recovery to the particular orientation that was fitted.

The same map has different effects across channel states, since it substantially improves the history-only and full-update states while worsening the pre-update state, so that in this event its benefit is concentrated at the states that contain the history displacement. The factorial interaction becomes less negative during rescue, which means that the improvement coexists with a reduction in the beneficial interaction between the channels (Figure~\ref{fig:frame}).

The norms, the covariance spectrum, and the persistence intervals remain essentially unchanged, as expected for a common orthogonal map, while the mean squared distance to the earlier held-out activations increases slightly (Table~\ref{tab:writing_rescue_details}). Substantial behavioural recovery therefore occurs while the intrinsic geometry is preserved and this average distance to the earlier representation grows, so the functional change follows the new orientation that is presented to the fixed downstream network.

Across the other OLMo states, the effects range from partial recovery of the loss to deterioration, and improvements in loss sometimes leave accuracy unchanged, while at one site a mean shift even exceeds the fitted rotation (Table~\ref{tab:frame_all}). This variation makes the parameter state and the selected layer central to the outcome.

\paragraph{Effect sizes across models and states.}
The fitted rotations for Pythia produce much smaller endpoint changes across the tested sites, and seeds 12 and 13 of the text study show mostly small increases together with a few small improvements (Table~\ref{tab:rotation_replication}). The contrast with OLMo places the large rescue at a particular combination of model, task, state, and fitting procedure, and identifying which of these features makes recovery possible is the next question that this intervention raises.

\subsection{Sequential association learning}
\label{app:movedregistry}
Each task is a registry that assigns a distribution over two single-token answers, \textrm{red} and \textrm{blue}, to each of 32 entities, with target probabilities of $(0.9,0.1)$ or $(0.1,0.9)$ sampled separately for each registry from the data seed. Each entity appears in two training templates, one validation template, and one test template, and while the prompt forms are disjoint across these splits, the entity identities are shared, so that test performance measures how the learned associations are expressed in new prompt contexts.

For a prompt $x$, target distribution $q_x$, and task population $\mathcal D_T$, we use
\begin{equation}
 \ell_x(\theta)=\sum_{y\in\{\textrm{red},\textrm{blue}\}}q_x(y)
 \log\frac{q_x(y)}{p_\theta(y\mid x)},\qquad
 L_T(\theta)=\frac{1}{|\mathcal D_T|}\sum_{x\in\mathcal D_T}\ell_x(\theta),
 \label{eq:loss}
\end{equation}
where the model probabilities are normalised over the full vocabulary, so that the loss penalises both the probability assigned outside the two answers and an incorrect distribution between them. Registry losses are KL divergences in nats, whereas the one-hot text-classification losses introduced below are negative log-likelihoods (NLLs), also in nats. Two-answer accuracy selects the more probable of the two designated answers and compares it with the majority target.

We evaluate OLMo-1B \citep{groeneveld2024olmo}, Pythia-70M and Pythia-410M \citep{biderman2023pythia}, Qwen2.5-0.5B base and Instruct \citep{qwen2024qwen25}, SmolLM2-360M base and Instruct \citep{benallal2025smollm2}, and Qwen3-0.6B-Base \citep{qwen2025qwen3}, with three data and minibatch seeds per checkpoint and 128 task-$B$ updates per seed, which gives 24 trajectories and 3,072 second-task updates in total. All checkpoints, including the Instruct variants, receive the same plain-text prompts without a chat template.

\paragraph{Training and evaluation.}
We train all parameters for 600 task-$A$ updates followed by 128 task-$B$ updates, using Adam with a constant learning rate of $2\times10^{-5}$, $(\beta_1,\beta_2)=(0.9,0.99)$, $\epsilon=10^{-8}$, no weight decay, and global gradient clipping at norm 1. Each update accumulates two microbatches of two prompts sampled with replacement, and the optimiser moments and step counter are preserved at the transition between tasks, so that the optimiser history, initialised at the start of training, accumulates across both tasks. Training uses FP32 parameters with dropout disabled, while the final-logit log-softmax and the KL evaluation use FP64, and Appendix~\ref{app:reproduce} gives the model revisions and further numerical settings.

We distinguish stepwise deterioration, $\Delta L_{A,t}=L_A(\theta_{t+1})-L_A(\theta_t)$, from the loss of retention relative to the post-$A$ anchor, and we call an update a large local event when $\Delta L_{A,t}>0.05$, a threshold shared by all eight checkpoints. A sign reversal additionally requires $\nabla L_A(\theta_t)^\top\delta_t<0$, where $\delta_t=\theta_{t+1}-\theta_t$, and the event counts obtained with a strict-zero threshold, which we use for one OLMo diagnostic, are identified separately.

\subsection{Frozen prediction across checkpoints}
\label{app:movedprediction}
\label{sec:prediction}
We fit logistic predictors on OLMo and reuse their feature definitions, normalisation, and coefficients on the other checkpoints. The loss-only predictor uses the current $A$ and $B$ losses, their previous changes, and the logarithm of the update index, and the loss-plus-projection predictor adds $\operatorname{asinh}(g_A^\top(h+c))$. The static geometry model further includes vector norms, pairwise dot products and cosines, and projections within three groups of layers, and the largest model adds recent changes in the vectors and their dot products, so that the four models contain 5, 6, 25, and 45 features, respectively (Appendix~\ref{app:predictors}).

For a target update $t$, lead 0 uses the features recorded immediately before that update, while lead 1 uses the features recorded immediately before update $t-1$ and is therefore the advance-warning setting. Examples that lack the preceding vector record are excluded, which leaves 127 and 126 examples per seed for the two leads, and the label is one if $\Delta L_{A,t}>0.05$ and zero otherwise. We report equal-weight means over seeds of AUROC, average precision (AP), and Brier score. The OLMo leave-one-seed-out scores are exploratory, because the protocol was developed on OLMo, whereas the trajectories of the target models never enter the fitting of the predictors. The diagnostic inputs include old-task gradients on the test population, which require access to retained evaluation data and additional backward passes.

\paragraph{Forecast timing and persistence baselines.}
We also evaluate raw scores on a completed-update decision timeline, which includes the most recent completed loss change. Immediately before update $d$, the previous projection is $g_{A,d-1}^{\top}\delta_{d-1}$ and the latest-loss baseline is the observed change in validation loss from update $d-1$, so both scores forecast the change in test loss at $d$ from information available after the preceding update. After the common warm-up period for history, there are 119 eligible forecasts per 128-update trajectory, and we also report horizons $d+1,d+3,d+7$ and event thresholds of 0.02, 0.05, and 0.10. The orientation of each score is fixed throughout the evaluation, AUROC and AP assess the raw rankings, and Brier scores apply only to the fitted probability predictors. The registry predictors instead use a feature timestamp that precedes the completion of the previous update.

\section{Reproducibility}
\label{app:reproduce}
\paragraph{Models and training configurations.}
The analysis uses recorded natural trajectories from OLMo, Pythia, Qwen2.5, SmolLM2, and Qwen3, and every checkpoint in the registry suite has three seeds, 600 acquisition updates, and 128 second-task updates under the configuration described in Section~\ref{sec:setting}. Acquisition uses a fixed budget of updates, and the training code also records whether the final $A$ validation KL is at most 0.25, a criterion that every run for which it is recorded passes, with their acquisition quality reported separately. Every measured step is evaluated on all 32 test prompts per task, and these full-population measurements are independent of the smaller routine evaluation setting in the source configuration.

\begin{table}[htbp]\centering\normalsize
\caption{Model repositories and exact revisions used in the experiments, where the revision identifiers specify the recorded repository versions. The association-learning suite uses SmolLM2-360M, and the natural-text studies use SmolLM2-135M.}\label{tab:checkpoints}
\begin{tabular}{p{0.43\linewidth}p{0.51\linewidth}}\toprule
Repository & Revision identifier\\\midrule
\nolinkurl{allenai/OLMo-1B-hf} & \texttt{aee7752d9c08ee4775e9\allowbreak b0091426d8410e8f6a89}\\
\nolinkurl{EleutherAI/pythia-70m} & \texttt{a39f36b100fe8a537781\allowbreak 0d56c3f4789b9c53ac42}\\
\nolinkurl{EleutherAI/pythia-410m} & \texttt{9879c9b5f8bea9051dcb\allowbreak 0e68dff21493d67e9d4f}\\
\nolinkurl{Qwen/Qwen2.5-0.5B} & \texttt{060db6499f32faf8b984\allowbreak 77b0a26969ef7d8b9987}\\
\nolinkurl{Qwen/Qwen2.5-0.5B-Instruct} & \texttt{7ae557604adf67be5041\allowbreak 7f59c2c2f167def9a775}\\
\nolinkurl{HuggingFaceTB/SmolLM2-360M} & \texttt{f8027fd0eaeea54caa13\allowbreak c31d31b9fdc459c38b49}\\
\nolinkurl{HuggingFaceTB/SmolLM2-360M-Instruct} & \texttt{a10cc1512eabd3dde888\allowbreak 204e902eca88bddb4951}\\
\nolinkurl{Qwen/Qwen3-0.6B-Base} & \texttt{da87bfb608c14b7cf20b\allowbreak a1ce41287e8de496c0cd}\\
\nolinkurl{HuggingFaceTB/SmolLM2-135M} & \texttt{93efa2f097d58c2a74874\allowbreak c7e644dbc9b0cee75a2}\\\bottomrule\end{tabular}\end{table}

Training disables dropout, TF32, and flash or memory-efficient attention, with OLMo using the math implementation of scaled dot-product attention and the other models using eager attention. Parameters, optimiser states, the forward computation, and parameter gradients all use FP32, while the log-softmax, the KL reductions, and the accumulation of vector dot products use higher precision where the implementation specifies it. The Qwen3 check records PyTorch 2.11.0+cu128 and Transformers 4.57.6, a maximum forward-pass discrepancy of $5.38\times10^{-6}$, and finite gradients for all unique parameters. All experiments ran on a GH200 host, and the reported results reflect these finite-precision computations.

Table~\ref{tab:datasetrevisions} lists the dataset versions recorded for the natural-history text study.

\begin{table}[htbp]\centering\normalsize
\caption{Dataset revisions for the natural-history text study.}\label{tab:datasetrevisions}
\begin{tabular}{p{0.20\linewidth}p{0.72\linewidth}}\toprule
Dataset & Revision identifier\\\midrule
SST-2 & \texttt{8d51e7e4887a4caaa95b3\allowbreak fbebbf53c0490b58bbb}\\
AG News & \texttt{eb185aade064a813bc0b7\allowbreak f42de02595523103ca4}\\
\bottomrule\end{tabular}\end{table}

\paragraph{Prompts and sampling.}
For registry $d\in\{A,B\}$ and a three-digit entity identifier $e$, training alternates between sampled examples from two templates:
\begin{quote}\normalsize
\texttt{Registry d. Object e. Category:}\\
\texttt{In registry d, the category of object e is:}
\end{quote}
The validation template is \texttt{Registry d lookup. Category for object e:}, and the test template is \texttt{Object e, registry d. Its category is:}, where the actual registry letter and entity identifier replace $d$ and $e$. Target assignments use Python's seeded random generator with seed $160001+s$, and microbatch sampling uses seed $7000001+100003s+101t$ for data seed $s$ and task-local update index $t$, with each microbatch sampled with replacement. Prompts are tokenised without added special tokens, and we check that the fixed answer strings \texttt{ red} and \texttt{ blue} each correspond to a single distinct token and that prefixes do not overlap across splits. Inputs are left-padded with explicit attention masks and position IDs, and prompts longer than 96 tokens are rejected.

\paragraph{Saved results and verification.}
The accompanying data summary records checkpoint configurations, per-seed trajectory and prediction metrics, selected path decompositions, derivative audits, and archive digests, and the figures summarise the path evaluations and factorial outcomes. For the two difficult OLMo updates, the reported path decompositions come from a targeted refinement on the same training data, and the refined curvature measurements are reported in their own table. The adapters for the native models are tested before training, and tests on tiny models check that the diagnostic instrumentation restores the natural trajectory of the parameters and the optimiser.

\section{Prediction methods and scores}
\label{app:predictors}
\paragraph{Feature sets.}
This section specifies the fitted registry predictor, while the raw-score analysis uses the completed-update timeline described in Section~\ref{sec:prediction}. All features are taken from the recorded pre-update state. The loss-only model uses $\log(1+t)$, $L_A$, $L_B$, and the difference between each current pre-update loss and its preceding pre-update value, and the loss-plus-projection model adds $\operatorname{asinh}(g_A^\top h+g_A^\top c)$, leaving out the explicitly recorded displacement residual, which is nevertheless retained in the path decomposition.

The 25-feature static geometry model contains these six features, four $\log(1+\|v\|)$ norms, six pairwise cosines, six $\operatorname{asinh}(u^\top v)$ dot products, and three old-task projections aggregated over successive thirds of the transformer layers. The layer-third aggregates cover only transformer layers, whereas the global quantities also include the embedding and readout blocks, and the static model retains the two preceding loss changes from the baseline, with undefined pairwise cosines encoded as zero.

The 45-feature model adds 20 temporal features, namely the consecutive cosine and the $\log(1+\text{relative change})$ of each of the four vectors and, for each of the pairs $h$--$g_A$, $c$--$g_A$, $h$--$g_B$, and $c$--$g_B$, the three terms of Equation~\ref{eq:temporal}, transformed with $\operatorname{asinh}$. Relative change is capped at $10^{12}$ before the transformation, and all of these features describe vectors in the native parameter space.

\paragraph{Fitting and timing.}
Each feature is standardised with the mean and standard deviation of the source fitting data, and standard deviations below $10^{-12}$ are replaced by one. Logistic regression minimises the summed binary log loss plus $\|w\|^2/(2C)$, with $C=0.1$ and an unpenalised intercept, using a damped Newton solver that checks convergence, and each lead has its own predictor. The exploratory OLMo scores leave an entire seed out of the fitting, while the transfer models are fitted on all three OLMo seeds and keep their feature normalisation and coefficients fixed, and the frozen artifacts in the Pythia, paired-model, and Qwen3 archives are identical.

For lead $k\in\{0,1\}$, target update $t$ is paired with the features at $t-k$, and since a usable feature record must have its preceding temporal record, lead 0 begins at target step 2 and lead 1 at target step 3. All loss changes used as features precede the feature measurement, so the design matrix uses only information available before the target. AUROC gives half credit for ties, AP groups tied scores, and the Brier score is the mean squared error of the predicted probability. Tables~\ref{tab:prediction_all} and~\ref{tab:prediction_seeds} report the scores without treating updates within a trajectory as independent statistical replicates.

\begin{table}[t]\centering\small
\caption{Transfer of the frozen registry predictor, compared with the specified fitted loss baseline. The comparator uses the pre-update feature timestamp, and the latest completed validation-loss baseline appears in Table~\ref{tab:text_timing}. All metrics are macro means across three seeds, the AP and Brier columns give loss-only / loss-plus-projection scores, and higher AUROC and AP and lower Brier are better.}\label{tab:prediction_main}
\setlength{\tabcolsep}{4pt}
\begin{tabular}{lrrrrr}\toprule
Target & Loss AUROC & +Proj. AUROC & Gain & AP & Brier\\\midrule
Pythia-70M & 0.721 & 0.845 & +0.124 & 0.380 / 0.563 & 0.084 / 0.073\\
Pythia-410M & 0.615 & 0.797 & +0.182 & 0.368 / 0.563 & 0.158 / 0.133\\
Qwen2.5 base & 0.686 & 0.822 & +0.136 & 0.397 / 0.519 & 0.149 / 0.121\\
Qwen2.5 Instruct & 0.667 & 0.804 & +0.137 & 0.352 / 0.520 & 0.137 / 0.120\\
SmolLM2 base & 0.861 & 0.957 & +0.096 & 0.491 / 0.575 & 0.075 / 0.066\\
SmolLM2 Instruct & 0.801 & 0.913 & +0.112 & 0.351 / 0.388 & 0.072 / 0.065\\
Qwen3 base & 0.759 & 0.866 & +0.108 & 0.450 / 0.650 & 0.114 / 0.097\\
\bottomrule\end{tabular}\end{table}

\begin{table}[htbp]\centering\small
\caption{Macro prediction metrics over three seeds each. OLMo uses exploratory leave-one-seed-out fits and all other checkpoints use frozen OLMo fits, with lead 0 serving as a current-update diagnostic and lead 1 as an advance warning.}\label{tab:prediction_all}
\setlength{\tabcolsep}{4pt}
\begin{tabular}{lrlrrr}\toprule
Checkpoint & Lead & Features & AUROC & AP & Brier\\\midrule
OLMo-1B & 0 & Loss & 0.843186 & 0.618288 & 0.127826\\
OLMo-1B & 0 & Loss + projection & 0.946781 & 0.858918 & 0.083301\\
OLMo-1B & 0 & Static geometry & 0.963795 & 0.860140 & 0.058668\\
OLMo-1B & 0 & Geometry + changes & 0.959504 & 0.870068 & 0.055377\\
OLMo-1B & 1 & Loss & 0.693560 & 0.344566 & 0.177768\\
OLMo-1B & 1 & Loss + projection & 0.792133 & 0.525299 & 0.149386\\
OLMo-1B & 1 & Static geometry & 0.794095 & 0.524619 & 0.141668\\
OLMo-1B & 1 & Geometry + changes & 0.802774 & 0.561423 & 0.135435\\
Pythia-70M & 0 & Loss & 0.913102 & 0.721913 & 0.061158\\
Pythia-70M & 0 & Loss + projection & 0.990916 & 0.915878 & 0.043341\\
Pythia-70M & 0 & Static geometry & 0.993164 & 0.942152 & 0.033262\\
Pythia-70M & 0 & Geometry + changes & 0.991746 & 0.901323 & 0.044954\\
Pythia-70M & 1 & Loss & 0.720669 & 0.380490 & 0.084033\\
Pythia-70M & 1 & Loss + projection & 0.844587 & 0.563012 & 0.073388\\
Pythia-70M & 1 & Static geometry & 0.749616 & 0.267869 & 0.071210\\
Pythia-70M & 1 & Geometry + changes & 0.761169 & 0.318977 & 0.073654\\
Pythia-410M & 0 & Loss & 0.817375 & 0.574221 & 0.127915\\
Pythia-410M & 0 & Loss + projection & 0.956281 & 0.899865 & 0.078635\\
Pythia-410M & 0 & Static geometry & 0.979097 & 0.958444 & 0.044161\\
Pythia-410M & 0 & Geometry + changes & 0.973979 & 0.937891 & 0.045473\\
Pythia-410M & 1 & Loss & 0.615346 & 0.368089 & 0.157556\\
Pythia-410M & 1 & Loss + projection & 0.797172 & 0.562751 & 0.133111\\
Pythia-410M & 1 & Static geometry & 0.810114 & 0.592330 & 0.124099\\
Pythia-410M & 1 & Geometry + changes & 0.795009 & 0.599889 & 0.120615\\
\bottomrule\end{tabular}\end{table}

\begin{table}[htbp]\centering\small
\caption{Macro prediction metrics over three seeds each. OLMo uses exploratory leave-one-seed-out fits and all other checkpoints use frozen OLMo fits, with lead 0 serving as a current-update diagnostic and lead 1 as an advance warning. (continued)}
\setlength{\tabcolsep}{4pt}
\begin{tabular}{lrlrrr}\toprule
Checkpoint & Lead & Features & AUROC & AP & Brier\\\midrule
Qwen2.5 base & 0 & Loss & 0.847762 & 0.557215 & 0.118395\\
Qwen2.5 base & 0 & Loss + projection & 0.986020 & 0.924421 & 0.070667\\
Qwen2.5 base & 0 & Static geometry & 0.985373 & 0.925991 & 0.046151\\
Qwen2.5 base & 0 & Geometry + changes & 0.980436 & 0.887451 & 0.049490\\
Qwen2.5 base & 1 & Loss & 0.685923 & 0.397295 & 0.149246\\
Qwen2.5 base & 1 & Loss + projection & 0.821588 & 0.518821 & 0.120706\\
Qwen2.5 base & 1 & Static geometry & 0.847979 & 0.527632 & 0.108072\\
Qwen2.5 base & 1 & Geometry + changes & 0.805287 & 0.479469 & 0.115605\\
Qwen2.5 Instruct & 0 & Loss & 0.862852 & 0.661355 & 0.112883\\
Qwen2.5 Instruct & 0 & Loss + projection & 0.973706 & 0.940259 & 0.074740\\
Qwen2.5 Instruct & 0 & Static geometry & 0.972882 & 0.939400 & 0.053557\\
Qwen2.5 Instruct & 0 & Geometry + changes & 0.974267 & 0.938311 & 0.052989\\
Qwen2.5 Instruct & 1 & Loss & 0.667438 & 0.352198 & 0.137408\\
Qwen2.5 Instruct & 1 & Loss + projection & 0.804089 & 0.520128 & 0.119725\\
Qwen2.5 Instruct & 1 & Static geometry & 0.807525 & 0.511516 & 0.112039\\
Qwen2.5 Instruct & 1 & Geometry + changes & 0.794633 & 0.478476 & 0.112517\\
SmolLM2 base & 0 & Loss & 0.962948 & 0.651521 & 0.057254\\
SmolLM2 base & 0 & Loss + projection & 0.997039 & 0.959023 & 0.042195\\
SmolLM2 base & 0 & Static geometry & 0.984982 & 0.820202 & 0.037652\\
SmolLM2 base & 0 & Geometry + changes & 0.995025 & 0.916600 & 0.026485\\
SmolLM2 base & 1 & Loss & 0.860798 & 0.490611 & 0.074893\\
SmolLM2 base & 1 & Loss + projection & 0.956794 & 0.574663 & 0.065872\\
SmolLM2 base & 1 & Static geometry & 0.821974 & 0.380200 & 0.111743\\
SmolLM2 base & 1 & Geometry + changes & 0.864977 & 0.394059 & 0.072251\\
\bottomrule\end{tabular}\end{table}

\begin{table}[htbp]\centering\small
\caption{Macro prediction metrics over three seeds each. OLMo uses exploratory leave-one-seed-out fits and all other checkpoints use frozen OLMo fits, with lead 0 serving as a current-update diagnostic and lead 1 as an advance warning. (continued)}
\setlength{\tabcolsep}{4pt}
\begin{tabular}{lrlrrr}\toprule
Checkpoint & Lead & Features & AUROC & AP & Brier\\\midrule
SmolLM2 Instruct & 0 & Loss & 0.961628 & 0.492285 & 0.054131\\
SmolLM2 Instruct & 0 & Loss + projection & 0.998380 & 0.980903 & 0.041752\\
SmolLM2 Instruct & 0 & Static geometry & 0.995311 & 0.950626 & 0.037380\\
SmolLM2 Instruct & 0 & Geometry + changes & 0.999145 & 0.989933 & 0.021764\\
SmolLM2 Instruct & 1 & Loss & 0.800813 & 0.351235 & 0.071948\\
SmolLM2 Instruct & 1 & Loss + projection & 0.912504 & 0.388456 & 0.064845\\
SmolLM2 Instruct & 1 & Static geometry & 0.821417 & 0.381051 & 0.114311\\
SmolLM2 Instruct & 1 & Geometry + changes & 0.831170 & 0.359942 & 0.067854\\
Qwen3 base & 0 & Loss & 0.879006 & 0.614299 & 0.097098\\
Qwen3 base & 0 & Loss + projection & 0.969782 & 0.873334 & 0.061580\\
Qwen3 base & 0 & Static geometry & 0.971743 & 0.831049 & 0.048666\\
Qwen3 base & 0 & Geometry + changes & 0.965200 & 0.772964 & 0.052514\\
Qwen3 base & 1 & Loss & 0.758672 & 0.449712 & 0.114382\\
Qwen3 base & 1 & Loss + projection & 0.866314 & 0.649698 & 0.097294\\
Qwen3 base & 1 & Static geometry & 0.813267 & 0.503120 & 0.105181\\
Qwen3 base & 1 & Geometry + changes & 0.827901 & 0.496144 & 0.098800\\
\bottomrule\end{tabular}\end{table}

\begin{table}[htbp]\centering\small
\caption{Per-seed advance-warning comparison. Paired entries give loss-only / loss-plus-projection scores, and N is the number of evaluated target updates.}\label{tab:prediction_seeds}
\setlength{\tabcolsep}{4pt}
\begin{tabular}{lrrrrrr}\toprule
Checkpoint & Seed & N & Events & AUROC & AP & Brier\\\midrule
OLMo-1B & 1 & 126 & 29 & 0.6715 / 0.7540 & 0.3827 / 0.5432 & 0.1711 / 0.1489\\
OLMo-1B & 2 & 126 & 14 & 0.6767 / 0.7621 & 0.1710 / 0.2542 & 0.1613 / 0.1368\\
OLMo-1B & 3 & 126 & 35 & 0.7325 / 0.8603 & 0.4800 / 0.7784 & 0.2009 / 0.1625\\
Pythia-70M & 1 & 126 & 7 & 0.4958 / 0.6579 & 0.0638 / 0.1474 & 0.0829 / 0.0743\\
Pythia-70M & 2 & 126 & 10 & 0.8241 / 0.9181 & 0.5467 / 0.7661 & 0.0835 / 0.0732\\
Pythia-70M & 3 & 126 & 13 & 0.8421 / 0.9578 & 0.5310 / 0.7755 & 0.0857 / 0.0727\\
Pythia-410M & 1 & 126 & 31 & 0.5379 / 0.7324 & 0.3513 / 0.5536 & 0.1899 / 0.1581\\
Pythia-410M & 2 & 126 & 16 & 0.6324 / 0.8614 & 0.2672 / 0.4909 & 0.1220 / 0.1041\\
Pythia-410M & 3 & 126 & 29 & 0.6758 / 0.7977 & 0.4857 / 0.6437 & 0.1608 / 0.1371\\
Qwen2.5 base & 1 & 126 & 21 & 0.5959 / 0.7710 & 0.2970 / 0.4541 & 0.1376 / 0.1182\\
Qwen2.5 base & 2 & 126 & 10 & 0.5966 / 0.7828 & 0.1351 / 0.2435 & 0.0995 / 0.0869\\
Qwen2.5 base & 3 & 126 & 44 & 0.8653 / 0.9110 & 0.7597 / 0.8589 & 0.2106 / 0.1570\\
Qwen2.5 Instruct & 1 & 126 & 27 & 0.6457 / 0.8081 & 0.4260 / 0.6032 & 0.1586 / 0.1372\\
Qwen2.5 Instruct & 2 & 126 & 11 & 0.6451 / 0.8126 & 0.1206 / 0.3093 & 0.1070 / 0.0916\\
Qwen2.5 Instruct & 3 & 126 & 26 & 0.7115 / 0.7915 & 0.5100 / 0.6478 & 0.1466 / 0.1304\\
SmolLM2 base & 1 & 126 & 5 & 0.7438 / 0.9421 & 0.2774 / 0.3558 & 0.0665 / 0.0600\\
SmolLM2 base & 2 & 126 & 3 & 0.9892 / 0.9919 & 0.5889 / 0.6389 & 0.0676 / 0.0599\\
SmolLM2 base & 3 & 126 & 16 & 0.8494 / 0.9364 & 0.6056 / 0.7293 & 0.0906 / 0.0778\\
SmolLM2 Instruct & 1 & 126 & 8 & 0.7246 / 0.8909 & 0.2948 / 0.3299 & 0.0750 / 0.0685\\
SmolLM2 Instruct & 2 & 126 & 1 & 0.7520 / 0.8880 & 0.0312 / 0.0667 & 0.0697 / 0.0612\\
SmolLM2 Instruct & 3 & 126 & 10 & 0.9259 / 0.9586 & 0.7276 / 0.7688 & 0.0711 / 0.0648\\
Qwen3 base & 1 & 126 & 14 & 0.6773 / 0.7857 & 0.4145 / 0.4885 & 0.1005 / 0.0909\\
Qwen3 base & 2 & 126 & 11 & 0.8593 / 0.9462 & 0.3951 / 0.7388 & 0.0911 / 0.0781\\
Qwen3 base & 3 & 126 & 28 & 0.7394 / 0.8670 & 0.5395 / 0.7218 & 0.1515 / 0.1229\\
\bottomrule\end{tabular}\end{table}

\section{Retention and selected events}
\label{app:events}
Table~\ref{tab:retention_seeds} reports all 24 trajectories under the common event threshold, and Table~\ref{tab:reversals} lists the four large reversals. The endpoint remainder of each reversal is computed directly from the measured loss and the initial slope, independently of whether a dense path is available, so the Pythia-410M and Qwen2.5 Instruct examples establish endpoint reversals from these measured quantities, while attribution to channels within the step is reported only for the separately audited paths.

\begin{table}[htbp]\centering\small
\caption{Per-seed task losses and event counts after 128 task-$B$ updates.}\label{tab:retention_seeds}
\setlength{\tabcolsep}{4pt}
\begin{tabular}{lrrrrrr}\toprule
Checkpoint & Seed & Anchor $L_A$ & Final $L_A$ & Final $L_B$ & Large & Reversals\\\midrule
OLMo-1B & 1 & 0.225999 & 0.844272 & 0.072626 & 29 & 2\\
OLMo-1B & 2 & 0.137821 & 0.447399 & 0.151627 & 14 & 0\\
OLMo-1B & 3 & 0.157457 & 0.627281 & 0.128464 & 35 & 0\\
Pythia-70M & 1 & 0.092410 & 0.602690 & 0.220334 & 7 & 0\\
Pythia-70M & 2 & 0.215931 & 0.515659 & 0.336593 & 12 & 0\\
Pythia-70M & 3 & 0.213590 & 0.418440 & 0.186810 & 13 & 0\\
Pythia-410M & 1 & 0.000343 & 1.000979 & 0.041929 & 31 & 0\\
Pythia-410M & 2 & 0.000154 & 0.554810 & 0.126810 & 16 & 0\\
Pythia-410M & 3 & 0.000778 & 0.807833 & 0.238539 & 29 & 1\\
Qwen2.5 base & 1 & 0.000163 & 0.894316 & 0.109290 & 21 & 0\\
Qwen2.5 base & 2 & 0.135829 & 0.554633 & 0.049059 & 10 & 0\\
Qwen2.5 base & 3 & 0.002182 & 1.159903 & 0.106209 & 44 & 0\\
Qwen2.5 Instruct & 1 & 0.001206 & 0.860381 & 0.139725 & 27 & 1\\
Qwen2.5 Instruct & 2 & 0.001794 & 0.530255 & 0.046934 & 11 & 0\\
Qwen2.5 Instruct & 3 & 0.002112 & 0.793077 & 0.091949 & 26 & 0\\
SmolLM2 base & 1 & 0.002965 & 0.938128 & 0.014385 & 5 & 0\\
SmolLM2 base & 2 & 0.007705 & 0.572143 & 0.012581 & 3 & 0\\
SmolLM2 base & 3 & 0.015339 & 0.942984 & 0.008793 & 16 & 0\\
SmolLM2 Instruct & 1 & 0.012660 & 0.936185 & 0.028321 & 8 & 0\\
SmolLM2 Instruct & 2 & 0.009741 & 0.621280 & 0.027070 & 1 & 0\\
SmolLM2 Instruct & 3 & 0.008117 & 0.708735 & 0.037097 & 10 & 0\\
Qwen3 base & 1 & 0.000460 & 0.457611 & 0.327182 & 14 & 0\\
Qwen3 base & 2 & 0.002415 & 0.712091 & 0.005682 & 11 & 0\\
Qwen3 base & 3 & 0.000162 & 0.560081 & 0.389287 & 28 & 0\\
\bottomrule\end{tabular}\end{table}

\begin{table}[t]\centering\small
\caption{All four large sign reversals on the old task in the measured trajectories. Only OLMo 1/21 has a corresponding audited dense path in this event set.}\label{tab:reversals}
\setlength{\tabcolsep}{4pt}
\begin{tabular}{llrrr}\toprule
Checkpoint & Seed/update & $\Delta L_A$ & $g_A^\top\delta$ & $R_A$\\\midrule
OLMo-1B & 1/21 & +0.287872 & -0.130264 & +0.418136\\
OLMo-1B & 1/76 & +0.064027 & -0.038522 & +0.102548\\
Pythia-410M & 3/18 & +0.051748 & -0.371233 & +0.422981\\
Qwen2.5 Instruct & 1/96 & +0.057541 & -0.056290 & +0.113831\\
\bottomrule\end{tabular}\end{table}

The selected Qwen3 paths distinguish the occurrence of a sign reversal from the size of its functional effect. At update 3, the change in old-task loss is $+0.000293927$, with slope $-0.000029029$ and remainder $+0.000322955$, whereas at update 59 the corresponding values are $+0.075012094$, $+0.065860746$, and $+0.009151348$, and task $B$ improves by 0.034693 at update 3 but worsens by 0.026703 at update 59. The magnitude, orientation, and joint terms for the old task are approximately $(-0.000020197,+0.000183901,+0.000159246)$ at update 3 and $(+0.027792165,-0.017319259,-0.001321733)$ at update 59. Since these terms decompose the remainder, adding them to the integrated history and current contributions would count part of the same loss change twice.

\section{Numerical integration and curvature audits}
\label{app:audits}
\paragraph{Path acceptance.}
The initial grid contains 17 fractions, and adaptive Simpson quadrature refines the intervals using both endpoint closure and nested estimates for the total directional derivative and for the history, current, residual, magnitude, orientation, and joint contributions. The configured absolute and relative integration tolerances are $10^{-5}$ and 0.005, with the absolute tolerance distributed across the width of each interval. Standard paths have a cap of 129 points and a depth limit of 10, which a targeted refinement of the difficult OLMo paths raises to 1025 and 16, and although the integration grids can differ between tasks, the underlying saved evaluations of the full gradient are shared.

The standard slope audits evaluate the fractions 0, 0.5, and 1 at finite-difference widths of 0.01 and 0.005, using central differences in the interior and one-sided differences at the endpoints, with an acceptance threshold of $0.0005+0.05\max(|s_{\rm FD}|,|s_{\rm AD}|)$. The refinements try additional widths down to 0.00015625 and require agreement between adjacent widths under the recorded refinement rule. We interpret a path only when both the quadrature and the slope criteria pass, and these checks concern mean quantities over the full population.

Of the 36 task-specific path decompositions, 32 pass both criteria, and the four that do not come from two updates, since OLMo seed 2/update 11 reaches its refinement cap without passing quadrature for either task, although it passes the slope checks and has small mean closure errors, while Pythia-70M seed 1/update 89 passes quadrature but fails the slope checks for both tasks. The corresponding integrated attributions are therefore excluded from the mechanistic examples in the paper. All 16 paired-model task paths and all four Qwen3 task paths pass, including all 96 and 24 of their individual slope checks, respectively, and Table~\ref{tab:path_audits} lists the failed results alongside the successful ones.

\begin{table}[htbp]\centering\small
\caption{Path checks, using the refined OLMo records for updates 1/21 and 2/11. The checks concern task-population means, and acceptance requires endpoint closure, component convergence, and slope agreement.}\label{tab:path_audits}
\setlength{\tabcolsep}{4pt}
\begin{tabular}{lllrrlr}\toprule
Checkpoint & Seed/update & Task & Points & Quad. & Slope & Closure\\\midrule
OLMo-1B & 1/021 & A & 157 & Pass & Pass & +3.82e-06\\
OLMo-1B & 1/021 & B & 217 & Pass & Pass & +7.48e-06\\
OLMo-1B & 1/048 & A & 33 & Pass & Pass & +1.70e-07\\
OLMo-1B & 1/048 & B & 33 & Pass & Pass & -9.20e-07\\
OLMo-1B & 2/011 & A & 1025 & Fail & Pass & +8.39e-07\\
OLMo-1B & 2/011 & B & 761 & Fail & Pass & +2.60e-05\\
OLMo-1B & 3/007 & A & 33 & Pass & Pass & -5.41e-08\\
OLMo-1B & 3/007 & B & 33 & Pass & Pass & +2.87e-07\\
Pythia-70M & 1/061 & A & 33 & Pass & Pass & -1.36e-05\\
Pythia-70M & 1/061 & B & 33 & Pass & Pass & -6.32e-08\\
Pythia-70M & 1/089 & A & 33 & Pass & Fail & -1.27e-05\\
Pythia-70M & 1/089 & B & 33 & Pass & Fail & +6.96e-06\\
Pythia-410M & 1/058 & A & 33 & Pass & Pass & +5.32e-07\\
Pythia-410M & 1/058 & B & 33 & Pass & Pass & -2.47e-07\\
Pythia-410M & 1/065 & A & 33 & Pass & Pass & -5.10e-08\\
Pythia-410M & 1/065 & B & 41 & Pass & Pass & -6.86e-08\\
Qwen2.5 base & 1/057 & A & 33 & Pass & Pass & -5.25e-07\\
Qwen2.5 base & 1/057 & B & 49 & Pass & Pass & -8.72e-07\\
Qwen2.5 base & 1/123 & A & 33 & Pass & Pass & +8.11e-08\\
Qwen2.5 base & 1/123 & B & 33 & Pass & Pass & -1.17e-07\\
Qwen2.5 Instruct & 1/034 & A & 33 & Pass & Pass & +2.24e-07\\
Qwen2.5 Instruct & 1/034 & B & 33 & Pass & Pass & -3.19e-07\\
Qwen2.5 Instruct & 1/107 & A & 33 & Pass & Pass & -8.07e-08\\
Qwen2.5 Instruct & 1/107 & B & 33 & Pass & Pass & +4.24e-08\\
\bottomrule\end{tabular}\end{table}

\begin{table}[htbp]\centering\small
\caption{Path checks, using the refined OLMo records for updates 1/21 and 2/11. The checks concern task-population means, and acceptance requires endpoint closure, component convergence, and slope agreement. (continued)}
\setlength{\tabcolsep}{4pt}
\begin{tabular}{lllrrlr}\toprule
Checkpoint & Seed/update & Task & Points & Quad. & Slope & Closure\\\midrule
SmolLM2 base & 1/028 & A & 33 & Pass & Pass & +4.60e-07\\
SmolLM2 base & 1/028 & B & 33 & Pass & Pass & +2.09e-07\\
SmolLM2 base & 1/073 & A & 33 & Pass & Pass & -2.44e-07\\
SmolLM2 base & 1/073 & B & 33 & Pass & Pass & -5.61e-08\\
SmolLM2 Instruct & 1/043 & A & 33 & Pass & Pass & -2.13e-07\\
SmolLM2 Instruct & 1/043 & B & 33 & Pass & Pass & +3.94e-07\\
SmolLM2 Instruct & 1/064 & A & 33 & Pass & Pass & -2.69e-07\\
SmolLM2 Instruct & 1/064 & B & 33 & Pass & Pass & -6.39e-08\\
Qwen3 base & 1/003 & A & 33 & Pass & Pass & -5.38e-09\\
Qwen3 base & 1/003 & B & 33 & Pass & Pass & +5.89e-07\\
Qwen3 base & 1/059 & A & 45 & Pass & Pass & -1.75e-07\\
Qwen3 base & 1/059 & B & 49 & Pass & Pass & -1.03e-07\\
\bottomrule\end{tabular}\end{table}

\paragraph{Curvature is a separate audit.}
The OLMo paths have eight curvature records across tasks and sites, all of which pass mixed symmetry, although only five pass the standard directional finite-difference check. The targeted refinement re-evaluates two locations in each difficult update, and all four refined measurements at seed 1/update 21 resolve, whereas at seed 2/update 11 both old-task measurements resolve but both new-task measurements fail the stability check across adjacent widths, so these remain unresolved even though the narrowest finite difference lies within the agreement tolerance. We report a refined measurement in place of the standard one only where the refined checks pass.

At the two refined sites of update 21, the old-task curvature is approximately $-127.073$ and $-127.871$, with history--history contractions of $-117.350$ and $-118.222$, while the new-task directional curvature at the same points is approximately $+20.523$ and $+27.130$. This local contrast is compatible with either sign of the full loss change, which depends on the entire path. The 16 paired-model and four Qwen3 sampled curvature records pass both the mixed-symmetry and the finite-difference checks.

\begin{table}[t]\centering\small
\caption{Refined OLMo curvature audits. A measurement passes when the finite differences agree and are stable across adjacent widths and mixed symmetry holds, and the width column gives the last attempted finite-difference width for each measurement.}\label{tab:curvature_refined}
\setlength{\tabcolsep}{4pt}
\begin{tabular}{lrlrllr}\toprule
Seed/update & $\alpha$ & Task & $\delta^\top H\delta$ & Sym. & FD & Last width\\\midrule
1/021 & 0.582031250 & A & -127.072809 & Pass & Pass & 0.00250000\\
1/021 & 0.582031250 & B & +20.523470 & Pass & Pass & 0.00500000\\
1/021 & 0.583984375 & A & -127.870972 & Pass & Pass & 0.00250000\\
1/021 & 0.583984375 & B & +27.130345 & Pass & Pass & 0.00250000\\
2/011 & 0.343383789 & A & +0.053302 & Pass & Pass & 0.00125000\\
2/011 & 0.343383789 & B & -46.336686 & Pass & Fail & 0.00015625\\
2/011 & 0.343275070 & A & +0.048840 & Pass & Pass & 0.00125000\\
2/011 & 0.343275070 & B & -46.414514 & Pass & Fail & 0.00015625\\
\bottomrule\end{tabular}\end{table}

\section{Activation intervention, controls, and topology}
\label{app:frame}
\paragraph{Fitting and held-out evaluation.}
The selected OLMo events are updates 20, 21, and 48 of seed 1 and update 7 of seed 3. Entity identities are partitioned into 16 calibration and 16 test identities with NumPy random-generator seed 8041, and calibration uses one validation prompt per task for each calibration entity, which gives 32 paired activation rows on which the fit minimises the activation discrepancy. For post-update rows $X$ and pre-update rows $Y$, we solve $\min_{Q^\top Q=I,\det Q=1}\|XQ-Y\|_F^2$ within their joint row span and assign the identity to directions outside that span. At the highlighted layer-0 event the rank is 64 in dimension 2048, and the fitted subspace rotation has an orthogonality error of $1.15\times10^{-14}$, while applying it in FP32 produces the small preservation errors reported in the main text.

The final-token residual output of one transformer block is patched for one prompt at a time, with all other token positions and all downstream weights held fixed, and all controls share the same fitting partition. The random orthogonal controls embed the same subspace rotation into four random orthonormal bases, which preserves its eigenangles while allowing the per-prompt displacement norm to vary, whereas the additive controls match that displacement norm separately for each prompt and channel state, so the two matching schemes answer different control questions.

The same map is applied to all four parameter corners at each fraction, which allows $J_T$ to be compared across corners, since refitting a separate map at each corner would change the intervention itself. The residual is assigned to the $uvr$ term so that corner 11 reconstructs the recorded update, and the factorial holds the denominator, the learning rate, and the channels fixed throughout the evaluation.

\paragraph{Full endpoint comparison.}
Table~\ref{tab:frame_all} includes every tested event and layer, and it shows that the large layer-0 rescue at update 21 is accompanied by smaller, absent, or harmful effects elsewhere, that loss can improve without any change in two-answer accuracy, and that at layer 9 of seed 3/update 7 the mean-shift control improves old-task loss slightly more than the fitted rotation. These observations limit how far the strongest selected case can be generalised.

\begin{table}[t]\centering\small
\caption{Fitted-rotation endpoint effects at $\alpha=1$ on 16 held-out entities. Loss changes compare patched with unpatched endpoints, accuracy entries are unpatched / patched percentages, the mean shift is the corresponding old-task control, and $\Delta J_A$ is a separate interaction contrast.}\label{tab:frame_all}
\setlength{\tabcolsep}{4pt}
\begin{tabular}{lrrrllr}\toprule
Seed/update & Layer & $\Delta L_A$ & $\Delta L_B$ & A acc. (\%) & Mean shift & $\Delta J_A$\\\midrule
1/020 & 0 & +0.044355 & -0.018944 & 100.0 / 100.0 & +0.043797 & +0.129258\\
1/020 & 1 & +0.002779 & +0.019591 & 100.0 / 100.0 & +0.050376 & +0.000436\\
1/020 & 9 & +0.078667 & +0.170582 & 100.0 / 93.8 & -0.009604 & +0.025977\\
1/021 & 0 & -0.266076 & -0.061470 & 50.0 / 100.0 & -0.047052 & +0.066994\\
1/021 & 1 & +0.019199 & +0.041272 & 50.0 / 50.0 & -0.199902 & +0.017846\\
1/021 & 9 & -0.014137 & +0.005740 & 50.0 / 50.0 & -0.041969 & -0.013997\\
1/048 & 0 & -0.196315 & -0.155932 & 50.0 / 50.0 & -0.087746 & +0.011014\\
1/048 & 1 & -0.132585 & -0.100554 & 50.0 / 50.0 & -0.021683 & +0.008021\\
1/048 & 9 & -0.194095 & -0.106920 & 50.0 / 50.0 & -0.097688 & +0.015086\\
3/007 & 0 & -0.015371 & -0.008960 & 31.2 / 31.2 & -0.013252 & +0.000242\\
3/007 & 1 & -0.029097 & -0.042132 & 31.2 / 31.2 & -0.039511 & -0.001687\\
3/007 & 9 & -0.336177 & -0.158858 & 31.2 / 31.2 & -0.345039 & +0.005324\\
\bottomrule\end{tabular}\end{table}

\paragraph{Geometric invariants.}
The point cloud contains the final-token activations of the 16 test entities for one task, with their identities preserved across conditions. Euclidean distances are divided by the median pairwise distance of the pre-update reference cloud, and the Vietoris--Rips filtration includes all vertices, edges, and triangles, with each simplex entering at its maximum edge length and homology reduced over $\mathrm{GF}(2)$. We retain the $H_0$ and $H_1$ intervals and also sample Betti curves at 31 thresholds from 0 to 3, although the full interval calculation extends beyond this sampled range.

For the highlighted event, the single finite $H_1$ interval changes from approximately $(0.821767186,0.877678021)$ to $(0.821766089,0.877676689)$, a maximum endpoint difference of $1.33\times10^{-6}$ under this matching, while sorted matching of the finite $H_0$ intervals gives a maximum difference of $1.05\times10^{-5}$, and the essential connected component is present in both conditions. These discrepancies bound the stated explicit matchings of intervals. The sampled Betti curves are unchanged, but a short $H_1$ interval lies between adjacent sampling thresholds, which is why we retain the full intervals.

In exact arithmetic, a common orthogonal map preserves within-cloud Euclidean distances exactly, so the measured discrepancies serve as a check of numerical preservation at the patched layer. The archived \texttt{distance\_relative\_change} field uses the pre-update cloud as its reference and is interpreted with that reference throughout, while functional rescue is measured separately, with the downstream computation held fixed as the representation is reoriented.

\section{Text-task forecasts and interventions}\label{app:extensions}
The Pythia text study examines sentiment classification followed by topic classification. In seed 11, the restricted-answer accuracy on the old task rises from 46.9\% before acquisition to 82.8\% after it, and then falls to 68.8\% during natural training on the second task, with a final full-vocabulary accuracy of 53.1\%, while the old-task NLL increases from 0.598 to 2.909, so acquired behaviour is also clearly impaired in the text setting. Seeds 12 and 13 reach acquired losses of 1.140 and 1.340, and all comparisons therefore use the acquired state within each seed as their reference.

The seed-11 study contains 36 conditions across the registry and text tasks, and the study of seeds 12 and 13 contains 27 full 128-update trajectories and 24 selected eight-update branches, including five random schedules for each tested action budget, with all conditions grouped within the stated training seeds. The frozen registry prediction coefficients were fitted on OLMo diagnostic populations, whereas the text studies use validation gradients, which is an explicit shift in measurement, and the latest-loss comparison is unfitted and uses the common decision timestamp described in Section~\ref{sec:prediction}.

\begin{table}[htbp]
\centering\small
\caption{Pythia text-task forecast baselines on the same 119 eligible decisions per seed. The target is an increase of more than 0.05 nats in $A$-test loss at the next update, and all entries are AUROC. The fitted loss-history baseline differs from the latest available change in validation loss.}\label{tab:text_timing}
\setlength{\tabcolsep}{3pt}
\begin{tabular}{lrrr}\toprule
Score & Seed 11 & Seed 12 & Seed 13\\\midrule
Previous actual projection & .8183 & .8465 & .9272\\
Previous task-relative cosine & .8081 & .8556 & .9274\\
Latest validation-loss increase & .8509 & .8522 & .9405\\
Frozen loss-history predictor & .7029 & .7117 & .8305\\
Previous population-gradient interference & .7551 & .6218 & .6777\\
Positive events & 56 & 58 & 56\\
\bottomrule\end{tabular}
\end{table}

\begin{table}[htbp]
\centering\small
\caption{Pythia text-policy comparison, where each cell gives the final $A/B$ NLL, so that the two task losses together describe the trade-off between retention and acquisition.}\label{tab:text_policies}
\setlength{\tabcolsep}{3pt}
\begin{tabular}{lrr}\toprule
Condition & Seed 12 & Seed 13\\\midrule
Natural & 4.518 / .999 & 1.959 / 1.050\\
Projection-triggered & 4.433 / .724 & 2.322 / 1.360\\
Loss-history-triggered & 2.047 / .728 & 2.028 / 1.529\\
Constant quarter learning rate & 2.575 / .990 & 1.929 / 1.090\\
Reset first moment, nominal rate & 4.277 / .785 & 2.434 / .929\\
Reset first moment, quarter rate & 2.190 / .891 & 1.877 / 1.137\\
\bottomrule\end{tabular}
\end{table}

The projection policy takes 16 actions in each of seeds 12 and 13 and the loss-history policy takes 14 and 16, while random schedules with matched action counts serve as retrospective controls for timing. The old-task loss under the projection policy lies within the random range in both seeds, $[3.485,7.185]$ and $[1.125,3.269]$. The fixed new-task validation targets and sustained crossing rules are retained in the accompanying numerical exports, and the relative outcomes of the policies vary across these comparisons, while the five control schedules describe the variability within a seed under a matched action budget.

\paragraph{Short matched-history continuations.}
All 156 global and blockwise norm-group checks pass in the selected branch study for seeds 12 and 13, with a maximum relative discrepancy of $1.36\times10^{-5}$, and the matching constrains only the first displacement. At seed 12/update 9, the block-matched reset branch improves both final losses after eight updates relative to preserved history, whereas at seed 13/update 6, preserved history immediately changes the $A/B$ loss by $-0.067790/-0.674917$, while the block-matched reset changes it by $+0.128456/+0.139535$. Only two of the six selected and fixed sites increase the full $A$-test loss at the natural first step, so the remaining four provide comparison sites at which the step improves the loss.

\paragraph{Path integrals and derivative checks.}
All twelve task-specific path integrals for seeds 12 and 13, together with their sampled directional-Hessian audits, pass their saved criteria. At seed 12/update 9, the change of $+0.454895$ in $A$ loss decomposes along the recorded path into $+0.402912$ from history and $+0.051983$ from the current channel, and at seed 13/update 6, history contributes $-0.087281$ and the current channel $+0.019484$, which gives $-0.067790$. These integrals describe the recorded fixed path, while the reset branches evaluate the alternative paths induced by the intervention.

For the task-$B$ curvature at seed 11/update 5, the sites $\alpha=.30$ and $.328125$ pass the criterion based on the two smallest widths, but $\alpha=.36$ remains unresolved, since the narrowest width alone is insufficient under the saved adjacent-width rule. The endpoint loss and the separately resolved path integral keep their recorded status, because the curvature audit concerns only the specified sampled locations.

\paragraph{Frame-intervention replication.}
The Pythia seed-11 suite probes six selected updates at two layers each, and the fitted effects are small, with the largest absolute change in endpoint task loss at about 0.000514. At twelve old-task event and layer sites in seeds 12 and 13, the fitted changes range from $-0.000156$ to $+0.002701$, with only three of them negative, so under the specified tasks and fitting rules the fitted effects at these replication sites are much smaller than the highlighted OLMo rescue, whose complete endpoint table remains in Appendix~\ref{app:frame}.

\subsection{Detailed text paths, history branches, and rotation replication}
The tables below give event-level path integrals, history interventions, and rotation effects. The path integrals use the stated task population, the branch matching constrains only the first displacement, after which each optimiser state evolves independently, and the rotation effects use the held-out intervention populations.

\begin{table}[htbp]
\centering\small
\caption{Selected Pythia text paths. Entries are changes in old-task test loss, or contributions to them, in nats, and all paths pass their recorded integration and sampled Hessian checks.}\label{tab:text_path_details}
\setlength{\tabcolsep}{3pt}
\begin{tabular}{lrrrrr}\toprule
Seed/update & $\Delta L_A$ & Initial slope & Remainder & History & Current\\\midrule
12/8 & +0.074220 & -0.012870 & +0.087090 & +0.052256 & +0.021964\\
12/9 & +0.454895 & --- & --- & +0.402912 & +0.051983\\
13/6 & -0.067790 & --- & --- & -0.087281 & +0.019484\\
\bottomrule\end{tabular}
\end{table}

\begin{table}[htbp]
\centering\small
\caption{Selected Pythia text branches. Reset rows match the initial displacement norm within blocks, and immediate changes and losses after eight branch updates are in nats.}\label{tab:text_branch_details}
\setlength{\tabcolsep}{3pt}
\begin{tabular}{llrrrr}\toprule
Seed/update & History & First $\Delta L_A$ & First $\Delta L_B$ & Final $L_A$ & Final $L_B$\\\midrule
12/9 & Preserved & +0.454895 & -0.086883 & 3.095976 & 2.553748\\
12/9 & Block-matched reset & +0.194391 & -0.248681 & 2.075273 & 2.046184\\
13/6 & Preserved & -0.067790 & -0.674917 & 1.404899 & 1.509092\\
13/6 & Block-matched reset & +0.128456 & +0.139535 & 1.130651 & 3.194577\\
\bottomrule\end{tabular}
\end{table}

\begin{table}[htbp]
\centering\small
\caption{Pythia rotation replication. Endpoint effects compare patched and unpatched losses on the intervention evaluation populations.}\label{tab:rotation_replication}
\setlength{\tabcolsep}{3pt}
\begin{tabular}{p{.33\linewidth}p{.61\linewidth}}\toprule
Evaluation & Recorded endpoint effect\\\midrule
Seed 11, tested conditions & Largest absolute task-loss effect $\approx5.14\times10^{-4}$ nats\\
Seeds 12/13, twelve old-task sites & Range $[-0.000156,+0.002701]$ nats; three improvements\\
\bottomrule\end{tabular}
\end{table}

\section{Optimiser calibration, confirmation, and numerical checks}
\label{app:optimizer}
The optimiser study comprises 62 runs, namely 24 primary calibration candidates, 24 primary confirmations, six paired-label confirmations, six appendix calibration candidates, and two appendix confirmations, and its design keeps the calibration of learning rates separate from the confirmation seeds. The 24 main confirmations cover two models, four optimisers, and three seeds, and one of them fails the acquisition gate for the old task, since SmolLM2 momentum SGD seed 22 has a validation loss improvement of 1.487090 but a restricted accuracy of only .375. All six label controls pass, whereas the Muon appendix confirmation also fails acquisition, so a count restricted to the main confirmations would miss this second failed run.

\paragraph{Development selection.}
Hashes of the raw texts allocate 20\% of each corpus to development and 80\% to confirmation. Calibration uses 300 acquisition and 64 second-task updates and evaluates only the training and validation populations for selection, while confirmation uses 600 and 128 updates and the appendix confirmations use 600 and 64. The acquisition gate requires a validation improvement of at least .1 nat and a restricted-answer accuracy of at least .6, and among the eligible candidates we select the one that minimises the mean $B$-validation NLL plus the positive excess of $A$-validation loss above the acquired anchor, with coefficient one and ties broken in favour of the smaller learning rate. Selection uses this fixed grid of candidates, and the dataset revisions, recorded in the companion evidence, are fixed throughout the suite.

Each model receives the SGD candidates $.001,.01,.1$, the momentum-SGD candidates $.0001,.001,.01$, and, separately for both Adam variants, the candidates $5\times10^{-6},2\times10^{-5},8\times10^{-5}$, and the selected rates are:
\begin{center}\small
\begin{tabular}{lrrrr}\toprule
Model & SGD & Momentum SGD & Adam $\beta_1=0$ & Adam\\\midrule
Pythia-410M & .01 & .01 & $5\times10^{-6}$ & $5\times10^{-6}$\\
SmolLM2-135M & .1 & .01 & $2\times10^{-5}$ & $8\times10^{-5}$\\\bottomrule
\end{tabular}
\end{center}
Momentum is .9, Adam uses $\beta_2=.99$, $\epsilon=10^{-8}$, and the stated first-moment setting, and SGD momentum is classical, without Nesterov acceleration or dampening. All methods use zero weight decay, global clipping at norm one, and identical minibatch schedules within a seed, and each method learns its own old-task anchor. SmolLM2 with $\beta_1=0$ shows favourable descriptive retention, but this comes together with a learning rate four times smaller than that of Adam and with different acquired anchors.

\begin{table}[t]\centering\small
\caption{All main optimiser confirmation trajectories, including the acquisition failure. Losses are full-vocabulary NLL in nats.}\label{tab:optimizer_all}
\setlength{\tabcolsep}{4pt}
\begin{tabular}{llrlrrrr}\toprule
Model & Optimiser & Seed & LR & Gate & $A_0$ & $A_f$ & $B_f$\\\midrule
Pythia-410M & Adam & 21 & 5e-06 & yes & 1.054 & 1.729 & 0.525\\
Pythia-410M & Adam $\beta_1=0$ & 21 & 5e-06 & yes & 1.173 & 2.065 & 0.369\\
Pythia-410M & Momentum SGD & 21 & 1e-02 & yes & 1.280 & 2.630 & 0.479\\
Pythia-410M & SGD & 21 & 1e-02 & yes & 0.701 & 1.377 & 0.475\\
Pythia-410M & Adam & 22 & 5e-06 & yes & 0.928 & 3.772 & 0.564\\
Pythia-410M & Adam $\beta_1=0$ & 22 & 5e-06 & yes & 1.192 & 2.316 & 0.634\\
Pythia-410M & Momentum SGD & 22 & 1e-02 & yes & 0.873 & 1.111 & 0.851\\
Pythia-410M & SGD & 22 & 1e-02 & yes & 0.679 & 3.252 & 0.819\\
Pythia-410M & Adam & 23 & 5e-06 & yes & 0.684 & 2.045 & 0.493\\
Pythia-410M & Adam $\beta_1=0$ & 23 & 5e-06 & yes & 1.099 & 2.555 & 0.766\\
Pythia-410M & Momentum SGD & 23 & 1e-02 & yes & 1.036 & 2.702 & 0.904\\
Pythia-410M & SGD & 23 & 1e-02 & yes & 0.881 & 0.630 & 0.500\\
SmolLM2-135M & Adam & 21 & 8e-05 & yes & 0.748 & 2.187 & 0.426\\
SmolLM2-135M & Adam $\beta_1=0$ & 21 & 2e-05 & yes & 0.701 & 0.782 & 0.402\\
SmolLM2-135M & Momentum SGD & 21 & 1e-02 & yes & 1.251 & 1.435 & 0.349\\
SmolLM2-135M & SGD & 21 & 1e-01 & yes & 1.078 & 1.655 & 0.682\\
SmolLM2-135M & Adam & 22 & 8e-05 & yes & 1.056 & 1.557 & 0.719\\
SmolLM2-135M & Adam $\beta_1=0$ & 22 & 2e-05 & yes & 0.865 & 0.505 & 0.434\\
SmolLM2-135M & Momentum SGD & 22 & 1e-02 & no & 0.724 & 2.019 & 1.487\\
SmolLM2-135M & SGD & 22 & 1e-01 & yes & 0.872 & 1.059 & 0.539\\
SmolLM2-135M & Adam & 23 & 8e-05 & yes & 0.963 & 1.499 & 0.411\\
SmolLM2-135M & Adam $\beta_1=0$ & 23 & 2e-05 & yes & 0.541 & 0.427 & 0.326\\
SmolLM2-135M & Momentum SGD & 23 & 1e-02 & yes & 0.878 & 2.166 & 0.331\\
SmolLM2-135M & SGD & 23 & 1e-01 & yes & 0.971 & 1.678 & 0.358\\
\bottomrule\end{tabular}
\end{table}

\paragraph{Timing and seed dependence.}
Across individual optimiser runs, the same-update projection AUROC for Pythia spans 0.939--0.995. Table~\ref{tab:optimizer_timing} excludes only the failed main acquisition from its seed means, while Table~\ref{tab:optimizer_all} retains every main run, and the same-update and lagged scores answer different questions and have different eligible examples. For SmolLM2 with $\beta_1=0$, the three forecast sequences contain 29, nine, and four events, and these low event counts, together with the dependence between updates within a trajectory, limit how far the results can be interpreted. The bootstrap intervals in the numerical exports describe block variation within trajectories, and the combined-feature comparison, reported in Appendix~\ref{app:followup}, improves on one fitted baseline without consistently surpassing the latest-loss score.

\begin{figure}[htbp]\centering
\includegraphics[width=\linewidth]{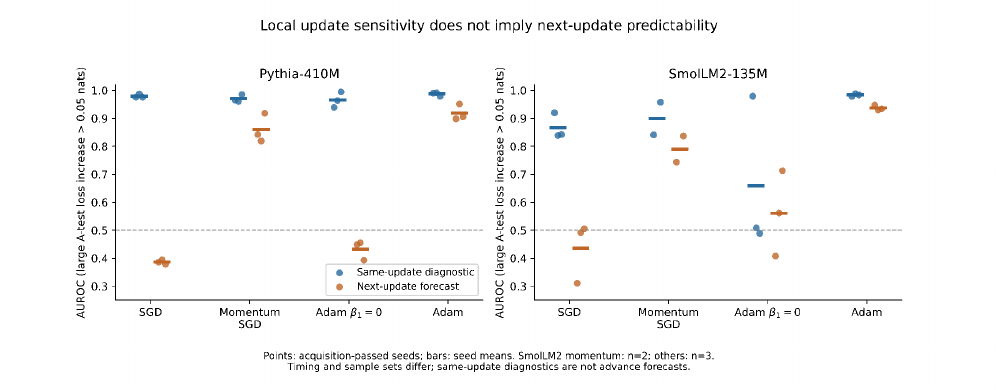}
\caption{Same-update projection diagnostics and preceding-update forecasts in the optimiser suite. Points are acquisition-passed seeds and horizontal bars are seed means, and the change between diagnostic and forecasting performance is especially clear for Pythia. The line at .5 marks chance AUROC for an uninformative ranking, and scores below chance retain their predeclared orientation.}
\label{fig:optimizer_geometry}
\end{figure}

\paragraph{Immediate norm-matched probes and paths.}
At the fixed updates 8 and 32, the reset probes preserve all optimiser state other than history, and global or blockwise scaling matches the norm of the actually assigned displacement to a relative tolerance of $10^{-3}$ and an absolute tolerance of $10^{-10}$. Across the main, label, and appendix conditions, all 120 recorded state audits and all 1,208 norm-group audits pass, including those of the failed-acquisition runs, since these audits check state preservation and norm matching rather than acquisition, and the maximum relative discrepancy in norm is $7.16\times10^{-4}$. These results are immediate evaluations, distinct from the eight-update Pythia continuations of Appendix~\ref{app:extensions}.

Twenty-four fixed $A$-validation paths use nine-point nested Simpson checks, and all of them pass the endpoint closure and the recorded quadrature-error criterion, while 22 pass the directional-Hessian audits, the two failures being Pythia seed 21 at Adam update 8 and at SGD update 32. The central differences use widths of .02, .01, and .005 and a tolerance of $2\times10^{-4}+.05\max(|a|,|b|)$, requiring the two smallest widths to agree both with autograd and with each other, and the curvature is sampled at the midpoint of the adjacent bracket with the largest change in slope, so that the Hessian checks evaluate those selected locations. None of these 24 audited fixed paths turns an initially favourable direction into a harmful endpoint, although the audit covers only the selected fixed steps.

\begin{table}[t]\centering\small
\caption{SmolLM2 paired answer-codebook controls, in which the acquired $A$ state is identical within each seed. Accuracy columns are percentages.}\label{tab:label_control}
\setlength{\tabcolsep}{4pt}
\begin{tabular}{rlrrrrr}\toprule
Seed & B labels & $A$ loss & Confusion & Leakage & Restricted & Full\\\midrule
21 & disjoint & 3.746 & 0.531 & 3.216 & 75.8 & 0.0\\
21 & shared & 1.470 & 1.120 & 0.350 & 57.8 & 53.9\\
22 & disjoint & 1.681 & 0.407 & 1.274 & 85.2 & 37.5\\
22 & shared & 1.367 & 0.722 & 0.645 & 70.3 & 43.8\\
23 & disjoint & 1.158 & 0.609 & 0.549 & 75.8 & 64.1\\
23 & shared & 1.801 & 0.571 & 1.230 & 71.1 & 23.4\\
\bottomrule\end{tabular}\end{table}

\paragraph{Label-control interpretation.}
The shared and disjoint pair preserves the old-task codebook and the acquired model within each seed, with paired underlying data and batch schedules, while the labels of the new task change together with its instructions. Since old-task leakage measures the total probability outside the old answer set, which can include the labels of the new task, comparing restricted and full-vocabulary accuracy shows how well the old answers compete both within their own set and against the rest of the vocabulary.

\paragraph{Appendix optimiser pilots.}
The blocked-Shampoo hybrid operates on hidden two-dimensional matrices, with $128\times128$ blocks, cumulative Gram statistics, damped inverse fourth roots, SGD-norm grafting, and momentum, while the remaining parameters use Adam. Its seed-31 confirmation acquires the old task and ends with an $A$ loss of .370598, down from .536864, and a $B$ loss of .548945 after 64 updates, which records the feasibility of this single-seed pilot.

The FP32-Muon hybrid uses an approximate Newton--Schulz matrix transform together with an auxiliary Adam group. Its seed-31 confirmation worsens the old-task validation loss by 2.639396 nats and reaches a restricted accuracy of .59375, failing the gate, so its retention measurements are reported together with this failed-acquisition status. Both pilots evaluate the specified custom hybrids under their limited tuning grids, and their full configurations and failures remain in the evidence exports.

\section{History reconstruction and cross-update geometry}
\label{app:historydynamics}
The Pythia and SmolLM2 history analysis covers 896 updates in seven trajectories, with Pythia contributing seed 51 with shared and disjoint answers and seed 52 with shared answers, and SmolLM2-135M contributing seeds 51 and 52 with both answer conditions. These are the first seven of the twelve natural-text trajectories in order of completion, and all seven satisfy the acquisition criterion for the old task. The archive reports that the exact endpoint replay and the algebraic checks pass for all updates and that 896 of 896 age-reconstruction checks pass, with a maximum relative residual of $9.074\times10^{-8}$ in the reconstruction of the first moment. These checks verify the saved replay and its algebraic reconstruction, while the derivative audits are reported with the selected finite paths.

\paragraph{Evaluation populations and loss decomposition.}
The replay uses all 32 $A$-validation examples and the first 32 examples in the fixed $A$-test order, an order that is independent of the observed harm. The source full-test evaluations use 128 examples, whereas the tables below use only the matched population of 32 examples on which both gradients and outcomes are computed. Confusion and leakage follow Equation~\ref{eq:answerdecomp}, and each row sums 128 local projections, each evaluated at its own pre-update parameter state. The remainder $E=\sum_t[\Delta L_t-g_t^\top h_t-g_t^\top c_t]$ includes the residual of the actual displacement and numerical discrepancies as well as finite-step effects, while the mixed history--current curvature, $h^\top Hc+c^\top Hh$, requires the separate path measurements.

\begin{table}[htbp]\centering\small
\caption{Decomposition of the total loss on the fixed 32-text test population. Values are in nats, and a negative observed change means a net improvement relative to the acquired anchor on this subset.}\label{tab:historytotal}
\setlength{\tabcolsep}{4pt}
\begin{tabular}{llrrrr}\toprule
Model & Seed / answers & Observed & History & Current & Remainder\\\midrule
Pythia-410M & 51 / disjoint & +4.519647 & +4.677210 & -0.894080 & +0.736518\\
Pythia-410M & 51 / shared & +1.423522 & +2.014214 & -1.046515 & +0.455823\\
Pythia-410M & 52 / shared & +0.428061 & +1.001943 & -1.301416 & +0.727534\\
SmolLM2-135M & 51 / disjoint & -0.432795 & -0.259566 & -0.288048 & +0.114819\\
SmolLM2-135M & 51 / shared & -0.460483 & -0.184538 & -0.528463 & +0.252518\\
SmolLM2-135M & 52 / disjoint & +1.490655 & +1.574926 & -0.353639 & +0.269367\\
SmolLM2-135M & 52 / shared & -0.121762 & +0.124659 & -0.490892 & +0.244471\\
\bottomrule\end{tabular}\end{table}

The corresponding decomposition of leakage by gradient age is given in Table~\ref{tab:historyages}.

\paragraph{Cross-time summary conventions.}
For each of the 127 pairs of adjacent updates in a trajectory, Equation~\ref{eq:symmetricpolar} gives the terms $O_t$, $G_t$, and $U_t$ for orientation, gradient magnitude, and channel magnitude, and we report
\begin{equation}
 f_O=\frac{\sum_t|O_t|}{\sum_t(|O_t|+|G_t|+|U_t|)},\qquad
 f_u=\frac{\sum_t|\bar g_t^\top\Delta u_t|}{\sum_t(|\bar g_t^\top\Delta u_t|+|\Delta g_t^\top\bar u_t|)},
\end{equation}
where absolute values are taken after summing the scalar contributions over parameters. Cosines below a norm floor of $10^{-12}$ are recorded as null, the comparisons between adjacent updates begin at the second update, and the global and layer-level records use the same definitions. The bin comparison in Table~\ref{tab:historyages} concerns cumulative loss projections, whereas these fractions concern their \emph{changes}, so the two quantities have different denominators.

\paragraph{Adaptive scaling.}
We write $u_t=K_tD_tN_t+\rho_t$, with $K_t=-\eta_t/(1-\beta_1^{k_t})$, $D_t=(\sqrt{v_t/(1-\beta_2^{k_t})}+\epsilon)^{-1}$ coordinatewise, $N_t=\beta_1m_{t-1}$ for history, and $N_t=(1-\beta_1)b_t$ for the current channel, where $\rho_t$ is the implementation residual. Then
\begin{equation}
 \Delta u_t=\overline{KD}_t\odot\Delta N_t
 +\bar N_t\odot(\bar K_t\Delta D_t+\bar D_t\Delta K_t)+\Delta\rho_t,
\end{equation}
and projecting each term onto $\bar g_t$ partitions the channel-motion term of Equation~\ref{eq:symmetricmotion}. Table~\ref{tab:historymotion} reports the sum of the absolute denominator terms divided by the sum of the absolute numerator and denominator terms. Changes in the clock are zero at the recorded precision in these records, while the implementation residuals are kept separately, so this ratio measures changes in the denominator around the recorded native preconditioner.

\begin{table}[htbp]\centering\small
\caption{Cross-time geometry of history for the total old-task loss. The first three columns give the defined absolute-contribution shares in percent, taken from separate or nested decompositions, and the last two columns give median cosines between consecutive updates. The shares follow the attribution conventions of Appendix~\ref{app:historydynamics}.}\label{tab:historymotion}
\setlength{\tabcolsep}{3.5pt}
\begin{tabular}{llrrrrr}\toprule
Model & Seed / answers & Orient. & Channel motion & Denom. & $\cos(h_t,h_{t-1})$ & $\cos(g_t,g_{t-1})$\\\midrule
Pythia-410M & 51 / disjoint & 77.19 & 65.71 & 3.21 & 0.9046 & 0.9984\\
Pythia-410M & 51 / shared & 83.69 & 75.56 & 1.82 & 0.9035 & 0.9983\\
Pythia-410M & 52 / shared & 78.57 & 73.35 & 2.62 & 0.8992 & 0.9980\\
SmolLM2-135M & 51 / disjoint & 82.77 & 68.88 & 1.80 & 0.9050 & 0.9989\\
SmolLM2-135M & 51 / shared & 82.02 & 68.03 & 2.12 & 0.8966 & 0.9978\\
SmolLM2-135M & 52 / disjoint & 77.35 & 70.41 & 2.62 & 0.8998 & 0.9993\\
SmolLM2-135M & 52 / shared & 77.82 & 73.41 & 1.80 & 0.9068 & 0.9983\\
\bottomrule\end{tabular}\end{table}

The corresponding global records for the current channel, the component-specific gradients, and the layer-level records are retained in the numerical supplement, and small net confusion projections may conceal cancellation across steps. The age table establishes a repeated signed pattern of leakage across the recorded bins, whereas tracking cohorts of fixed origin remains a separate analysis, and all of these estimates use the seven trajectories listed above.

\section{History interventions and forecasting}
\label{app:followup}
The matched-start intervention study uses Pythia-410M trajectories trained with Adam and SmolLM2-135M trajectories trained with Adam and with momentum SGD, for seeds 41--43, with the learning rates selected by the calibration in Appendix~\ref{app:optimizer}, namely $5\times10^{-6}$, $8\times10^{-5}$, and $.01$, respectively. Eight of the nine trajectories pass acquisition, while SmolLM2 momentum seed 43 fails and is excluded from the summaries of the acquired task. Each trajectory supplies common starting checkpoints at updates 8, 32, and 64, from which preserved history, reset history, and a reset with block-norm matching of the first displacement each continue for 32 updates. All 81 branches complete, all checks of the non-history state pass, and all 783 norm-group checks pass, with a maximum relative discrepancy of $5.812\times10^{-4}$, and these counts include the records of the failed-acquisition trajectory for transparency.

\paragraph{Continuation outcomes.}
Table~\ref{tab:followupbranches} averages the forks within each trajectory and then weights the acquired trajectories equally. Both Adam reset variants improve the final $A$ loss on average over forks in all six Adam model and seed trajectories, although the effects of individual forks and the mean retention vary, and SmolLM2 Adam seed 43, for instance, has a worse mean excess despite a better endpoint. Acquisition of the new task is assessed separately at common validation-loss targets, and at a common $B$-validation loss of $.75$ the mean block-matched difference in $A$ loss for Pythia is $+0.003580$ across nine pairs, while at a target of $1.0$ it is $-0.025831$, so the comparisons matched on acquisition remain mixed.

For Pythia, the mean benefit of the block-matched reset for the final old-task loss comes primarily from recovered leakage, since confusion worsens by $.099843$ nats while leakage improves by $.227013$, and full-vocabulary accuracy improves by 6.337 percentage points while restricted-answer accuracy changes by only 0.087 points. This intervention evidence is consistent with a functional distinction between discrimination and answer mass, while remaining separate from the attribution to age bins along natural trajectories.

\paragraph{Forecast comparisons.}
Under the completed-update decision timeline, the frozen combined-feature predictor fitted on pooled sources improves next-update AUROC over its frozen multivariate loss-history comparator in all eight acquisition-passed trajectories, with a mean difference of $.074943$ at the $.05$-nat threshold, and it improves the Brier score and the log loss in seven of the eight. The raw score given by the latest increase in validation loss, however, exceeds the combined predictor in seven of the eight trajectories (Table~\ref{tab:followupprediction}). The comparison between feature sets measures the joint contribution of the included features, and at horizon seven the mean difference in AUROC between the combined and loss-only models is $-.001966$. A retrospective subset of decisions preceded by quiet periods has only 4--9 positive events per trajectory and remains exploratory.

\begin{table}[htbp]\centering\small
\caption{Next-update AUROC for an increase of more than $.05$ nats in $A$-test loss, in the acquisition-passed trajectories of the matched-start study (seeds 41--43). ``Loss'' is the frozen multivariate loss-history model, and ``Latest'' is the raw latest increase in validation loss.}\label{tab:followupprediction}
\setlength{\tabcolsep}{4pt}
\begin{tabular}{llrrrrr}\toprule
Model & Optimiser & Seed & Loss & Combined & Previous proj. & Latest\\\midrule
Pythia-410M & Adam & 41 & .725313 & .837812 & .863750 & .871563\\
Pythia-410M & Adam & 42 & .862448 & .883289 & .911330 & .913225\\
Pythia-410M & Adam & 43 & .760782 & .828167 & .883423 & .883423\\
SmolLM2-135M & Adam & 41 & .887046 & .945764 & .931421 & .929628\\
SmolLM2-135M & Momentum & 41 & .575728 & .706349 & .720238 & .745370\\
SmolLM2-135M & Adam & 42 & .847882 & .892169 & .900193 & .903723\\
SmolLM2-135M & Momentum & 42 & .733963 & .842397 & .875611 & .885705\\
SmolLM2-135M & Adam & 43 & .792208 & .848966 & .871092 & .876864\\
\bottomrule\end{tabular}\end{table}

\paragraph{Paired finite paths.}
All 18 selected paths pass their saved criteria for slope closure and curvature integrals, but only 16 pass the derivative checks, since the preserved and block-reset paths of Pythia seed 41 at fork 8 remain unresolved, which leaves 10 of the 12 paired comparisons fully audited. At fork 8 of SmolLM2 Adam, the paired effect of the block reset is $+.001915=+.017169-.015254$, with an independently integrated Hessian effect of $-.015311$, and at fork 8 of momentum SGD the reset effect is $-.009049=-.002572-.006477$, with a Hessian effect of $-.006476$. These paths use the 32-text validation population and decompose the immediate branch step, whereas the full intervention outcome spans 32 updates. All twelve supplied paired decompositions keep the sign of their local prediction, and full numerical validity requires the derivative audit to pass in addition to endpoint closure.

\section{Event-level loss and intervention measurements}
\label{app:writing_details}
The following tables give event-level losses, channel integrals, and activation-intervention effects for the association-learning study, where losses are KL divergences in nats, path measurements use the full task population, and activation-intervention measurements use the entities held out from the fitting of the rotation. As an example of the feature-set comparison in Appendix~\ref{app:predictors}, SmolLM2 base has a mean AUROC of 0.822 with static geometry and of 0.957 with loss plus projection.

\begin{table}[htbp]
\centering\small
\caption{Behavioural retention and reversal counts. Large events use the common threshold $\Delta L_A>0.05$, and strict-positive events use $\Delta L_A>0$.}
\label{tab:writing_retention_details}
\begin{tabular}{lrr}
\toprule
Qwen3 seed & Pre-$B$ accuracy (\%) & Post-$B$ accuracy (\%)\\
\midrule
1 & 100 & 46.875\\
2 & 100 & 62.5\\
3 & 100 & 50\\
\midrule
Event population & Loss increases & Sign reversals\\
\midrule
All checkpoints, large events & 421 & 4\\
OLMo, strict-positive events & 200 & 17\\
\bottomrule
\end{tabular}

\smallskip
\parbox{0.96\linewidth}{\small Large reversals occur at OLMo seed 1 updates 21 and 76, Pythia-410M seed 3 update 18, and Qwen2.5 Instruct seed 1 update 96, and all 24 trajectories end above their post-$A$ loss anchor.}
\end{table}

\begin{table}[htbp]
\centering\small
\caption{The highlighted OLMo reversal at seed 1, update 21. The components of the remainder sum to $R_A$ up to rounding, and the curvature entries are local directional contractions at the stated location on the path.}
\label{tab:writing_path_details}
\begin{tabular}{lr}
\toprule
Quantity & Value\\
\midrule
Old-task loss before / after & $0.089042\;/\;0.376914$\\
Initial history projection $g_A^\top h$ & $-0.089216$\\
Initial current projection $g_A^\top c$ & $-0.041048$\\
Initial full slope & $-0.130264$\\
Observed old-task loss change & $+0.287872$\\
Finite-step remainder $R_A$ & $+0.418136$\\
Integrated history contribution & $+0.309796$\\
Integrated current contribution & $-0.021920$\\
Remainder: magnitude & $-0.200633$\\
Remainder: orientation & $+0.187092$\\
Remainder: joint & $+0.431681$\\
Sampled update fraction $\alpha$ & $0.58203125$\\
Old-task directional curvature at $\alpha$ & $\approx-127.073$\\
New-task directional curvature at $\alpha$ & $\approx+20.523$\\
\bottomrule
\end{tabular}
\end{table}

\begin{table}[htbp]
\centering\small
\caption{Contrasting Qwen3 events for seed 1. The orientation term is a component of the finite-step remainder of the old task.}
\label{tab:writing_qwen_details}
\begin{tabular}{lrr}
\toprule
Quantity & Update 59 & Update 3\\
\midrule
Old-task loss change & $+0.075012$ & $+0.000294$\\
Old-task integrated history & $+0.061498$ & $-0.000307$\\
Old-task integrated current & $+0.013514$ & $+0.000601$\\
History share of old-task increase & $\approx82\%$ & ---\\
New-task integrated history & $+0.033516$ & ---\\
New-task integrated current & $-0.006812$ & ---\\
Old-task remainder & $+0.009151$ & ---\\
Old-task orientation term & $-0.017319$ & ---\\
\bottomrule
\end{tabular}
\end{table}

\begin{table}[htbp]
\centering\small
\caption{The highlighted activation rescue at OLMo seed 1, update 21, layer 0, evaluated on 16 entities held out from fitting. Control effects are patched minus unpatched old-task loss at the full-update endpoint, and the geometric discrepancies compare the rotated and unrotated activation clouds.}
\label{tab:writing_rescue_details}
\begin{tabular}{lr}
\toprule
Quantity & Value\\
\midrule
Pre-update / unpatched endpoint loss & $0.085549\;/\;0.376039$\\
Rotated endpoint loss & $0.109962$\\
Fraction of event loss increase recovered & $91.6\%$\\
Two-answer accuracy: pre / post / rotated & $100\%\;/\;50\%\;/\;100\%$\\
Entities with improved loss & $13/16$\\
Median entity loss change & $-0.273151$\\
New-task loss change under rotation & $-0.061470$\\
Inverse rotation effect & $-0.008761$\\
Mean-shift effect & $-0.047052$\\
Scalar-scaling effect & $+0.001254$\\
Random orthogonal effects (four draws) & $[-0.004163,+0.016888]$\\
Matched additive effects (four draws) & All positive\\
History-only loss: unpatched / rotated & $0.457500\;/\;0.152914$\\
Current-only loss: unpatched & $0.076161$\\
No-update loss: unpatched / rotated & $0.085549\;/\;0.135893$\\
Interaction $J_A$: unpatched / rotated & $-0.072074\;/\;-0.005080$\\
Maximum norm discrepancy & $1.23\times10^{-6}$\\
Relative covariance-spectrum discrepancy & $1.86\times10^{-6}$\\
Finite $H_0$ endpoint matching bound & $1.05\times10^{-5}$\\
Finite $H_1$ endpoint matching bound & $1.33\times10^{-6}$\\
MSE to pre-update activations: unpatched / rotated & $0.006027\;/\;0.006240$\\
\bottomrule
\end{tabular}

\smallskip
\parbox{0.96\linewidth}{\small At the preceding update 20, the layer-0 rotation changes old-task loss by $+0.044355$, and the full endpoint comparison across events and layers is in Appendix~\ref{app:frame}. Persistence bounds are in normalised distance units.}
\end{table}

\section{Derivations and optimiser conventions}
\label{app:formalmechanism}
\subsection{Adam and momentum SGD}
For a clipped minibatch gradient $b_t$, Adam updates its moments by
\begin{align}
 m_t&=\beta_1m_{t-1}+(1-\beta_1)b_t,&
 v_t&=\beta_2v_{t-1}+(1-\beta_2)b_t^{\odot2},\\
 P_t&=\operatorname{diag}\!\left[
 \frac{\eta_t}{(1-\beta_1^{k_t})(\sqrt{v_t/(1-\beta_2^{k_t})}+\epsilon)}\right],
\end{align}
where $k_t$ is the optimiser step counter, which includes the training on task $A$. Substituting the first recurrence into $-P_tm_t$ gives Equation~\ref{eq:channels}, and the difference from the assigned parameter displacement is kept in $r_t$, so that the initial old-task projection is
\begin{equation}
 p_{A,t}=g_{A,t}^{\top}\delta_t
 =g_{A,t}^{\top}h_t+g_{A,t}^{\top}c_t+g_{A,t}^{\top}r_t.
 \label{eq:projection}
\end{equation}
For classical momentum SGD, $z_t=\mu z_{t-1}+b_t$, so that $h_t=-\eta_t\mu z_{t-1}$ and $c_t=-\eta_t b_t$, while plain SGD has $h_t=0$, and Adam with $\beta_1=0$ removes the memory of the first moment while retaining the adaptation of the second. All angular quantities are computed in the native parameter coordinates of the model.

\subsection{The straight-path identity and its curvature remainder}
Let $f(\alpha)=L_T(\theta_t+\alpha\delta_t)$, so that $f'(\alpha)=\nabla L_T(\theta_t+\alpha\delta_t)^\top\delta_t$, and integrating from zero to one gives Equation~\ref{eq:path}, while linearity in $\delta_t=h_t+c_t+r_t$ gives
\begin{equation}
 I_T^h=\int_0^1\nabla L_T(\theta_t(\alpha))^\top h_t\,\dd\alpha,
 \quad I_T^c=\int_0^1\nabla L_T(\theta_t(\alpha))^\top c_t\,\dd\alpha,
 \quad\Delta L_T=I_T^h+I_T^c+I_T^r.
 \label{eq:integrated}
\end{equation}
If $L_T$ is twice continuously differentiable along the segment, integration by parts yields
\[
 \int_0^1(1-\alpha)f''(\alpha)\,\dd\alpha
 =-f'(0)+f(1)-f(0)=R_T,
\]
and since $f''(\alpha)=\delta_t^\top H_T(\theta_t(\alpha))\delta_t$, this proves Equation~\ref{eq:curvature}. Expanding its quadratic form gives Equation~\ref{eq:curvaturesplit}, with
\[
 R_T^{\rm res}=2r_t^\top\mathcal K_T(h_t+c_t)+r_t^\top\mathcal K_T r_t
\]
when the Hessian is symmetric, a property that the numerical measurements check independently. A favourable-to-harmful reversal requires $g_T^\top\delta_t<0$ and $R_T>-g_T^\top\delta_t$, and positive curvature at a single sampled point is not sufficient to determine this weighted integral.

\subsection{Magnitude and orientation along a finite step}
Write $D=\|\delta_t\|$, $n(\alpha)=\|\nabla L_T(\theta_t(\alpha))\|$, and $q(\alpha)=\cos(\nabla L_T(\theta_t(\alpha)),\delta_t)$. Since $f'(\alpha)=Dn(\alpha)q(\alpha)$, expanding around $n_0=n(0)$ and $q_0=q(0)$ gives
\begin{equation}
 R_T=D\int_0^1\left[(n-n_0)q_0+n_0(q-q_0)+(n-n_0)(q-q_0)\right]\dd\alpha,
 \label{eq:polar}
\end{equation}
where the three terms measure the change in gradient magnitude, the change in orientation, and their interaction, and together they sum to the same remainder as the curvature integral. The convention for the cosine at vanishing norms is specified in Appendix~\ref{app:protocols}.

\subsection{Symmetric cross-update attribution}\label{sec:historymath}
For any scalar sequences $x_t,y_t$,
\[
 \Delta(xy)_t=\bar x_t\Delta y_t+\bar y_t\Delta x_t,
\]
and applying this identity coordinatewise to $g_t^\top u_t$ and summing proves Equation~\ref{eq:symmetricmotion}, while applying it first to $(n_td_t)z_t$ and then to $n_td_t$ gives
\[
 \Delta(ndz)_t=\overline{nd}_t\Delta z_t
 +\bar z_t\bigl(\bar n_t\Delta d_t+\bar d_t\Delta n_t\bigr),
\]
which is Equation~\ref{eq:symmetricpolar}. Signed sums telescope, which is why we use the absolute-contribution normalisation of Appendix~\ref{app:historydynamics} to measure variation across updates.

Expanding $m_{t-1}$ from the task boundary gives
\[
 m_{t-1}=\beta_1^{t-1}m_A+(1-\beta_1)\sum_{j=1}^{t-1}\beta_1^{t-1-j}b_j,
\]
and multiplying by $-P_t\beta_1$ proves Equation~\ref{eq:historyages}. Every contribution is evaluated with the current $P_t$, which preserves their sum as the history channel, and grouping these terms by age produces the bins of Table~\ref{tab:historyages}.

\section{Detailed measurement protocols}\label{app:protocols}
\label{app:detailedmethods}
\subsection{Sequential association learning}
The registry suite uses 32 entities, each assigned a soft distribution over two designated single-token answers, and disjoint training, validation, and test templates probe the same identities. Eight checkpoints from the OLMo, Pythia, Qwen2.5, SmolLM2, and Qwen3 families, including paired base and Instruct variants, each contribute three seeds and 128 second-task updates, for a total of 3,072 updates. The models undergo 600 task-$A$ updates before task $B$, with the native Adam moments preserved at the transition, full-vocabulary KL is measured in nats, and the common threshold for a large event is an increase in old-task loss above $.05$. Appendix~\ref{app:movedregistry} gives the definitions of the loss, acquisition, and sampling, and these registry measurements are separate from the text-history results below.

\subsection{Text classification, optimiser controls, and functional outcomes}
\label{sec:textmethods}
The text setting uses SST-2 sentiment classification \citep{socher2013sentiment} followed by AG News topic classification \citep{zhang2015text}, both expressed as single-token answers to plain-text instructions. Each task has 1,024 training, 32 validation, and 128 test texts, with prefixes truncated to at most 256 tokens, and a partition based on text hashes separates the development and confirmation pools, within which examples can overlap across seeds. Operational measurements use gradients over the full validation population together with held-out test outcomes, which differs from the diagnostic gradients on the test population used in the registry study.

An initial Pythia-410M study uses seeds 11--13 to compare natural training, step reduction triggered by the projection or by the loss history, random schedules with matched action counts, lower constant learning rates, and resets of the first moment. Selected events also receive globally or blockwise norm-matched resets with short continuations, and we evaluate the quality of prediction and the quality of prevention separately, always reporting new-task acquisition alongside retention.

The optimiser study uses SmolLM2-135M and Pythia-410M, each trained with SGD, momentum SGD, Adam with $\beta_1=0$, and ordinary Adam, and each model--optimiser pair receives three development candidates for the learning rate. Eligible candidates improve the $A$-validation loss by at least 0.1 nat and reach a restricted-answer accuracy of 0.6, and selection minimises the mean $B$-validation loss plus the positive excess of the $A$-validation loss above its acquisition anchor. Confirmation uses seeds 21--23, and each optimiser trains both tasks with its native state, which yields learning rates and acquired $A$ anchors specific to each configuration. Paired fixed-step probes at updates 8 and 32 start from identical weights and optimiser state and change the first-moment history and, in the matched controls, the global or blockwise scale of the first displacement, while the second moment and the clock are retained, so that these probes measure the immediate effect of the modified update.

To separate confusion among answers from displacement of probability, let $S_T$ be the designated answer set of a task. For a one-hot answer $y\in S_T$, Equation~\ref{eq:answerdecomp} separates the full-vocabulary NLL into a conditional NLL and a leakage term, and for soft registry targets the first term becomes a conditional KL divergence. We report restricted-answer and full-vocabulary accuracy separately. The paired SmolLM2 label controls keep the old-task codebook and the acquired state identical while varying whether the answer tokens of the new task overlap with it, and Appendix~\ref{app:optimizer} gives all configurations, gate failures, and audit boundaries.

\subsection{Natural component and history-age replay}
\label{sec:historymethods}
We analyse seven native-Adam text trajectories, consisting of Pythia-410M seed 51 with shared and disjoint answer sets, Pythia-410M seed 52 with shared answers, and SmolLM2-135M seeds 51 and 52 under both conditions, each with 128 second-task updates, for 896 measured updates in total, and all of them pass the acquisition gate. The natural-text study contains twelve trajectories, and the history replay covers the first seven in order of completion, so the sample is unbalanced and includes paired conditions, with dependence both between updates and between paired trajectories.

The task protocol is the text setting above, with 600 acquisition updates and native Adam retained across the transition. The learning rates are $5\times10^{-6}$ for Pythia and $8\times10^{-5}$ for SmolLM2, taken from the calibration in Appendix~\ref{app:optimizer}, with $\beta_1=.9$, $\beta_2=.99$, $\epsilon=10^{-8}$, clipping at norm one, and no weight decay. At every replayed update, we compute full component gradients on all 32 validation texts and on a fixed subset consisting of the first 32 test texts, which differs from the 128-text test evaluation of the source trajectory, and the tables identify the population explicitly. Every acquired anchor and recorded parameter endpoint must replay with an identical hash, and the algebraic reconstruction is audited independently of the finite-difference and curvature checks on the selected paths, which are what assess numerical differentiability.

The seeds-41--43 intervention study contains nine natural trajectories, 81 matched-state branches, and a frozen comparison of combined-feature predictions, and its intervention and prediction populations are kept separate from the natural age analysis (Appendix~\ref{app:followup}). The Qwen3 registry evidence belongs to the association-learning suite, while the Qwen3 text and history experiments are documented in Appendix~\ref{app:newreplication}.

\subsection{Recording and auditing update geometry}
For the registry diagnostics, we compute the full test-population gradients $g_{A,t}$ and $g_{B,t}$ over every unique trainable parameter, without clipping or sketching, and for the four vectors $[h_t,c_t,g_{A,t},g_{B,t}]$ we record the Gram matrix, the norms, and the defined pairwise cosines, both globally and within disjoint blocks of parameter layers.

For the features of the registry predictor, we record the exact decomposition
\begin{equation}
 u_t^\top v_t-u_{t-1}^\top v_{t-1}
 =\Delta u_t^\top v_{t-1}+u_{t-1}^\top\Delta v_t
 +\Delta u_t^\top\Delta v_t,
 \label{eq:temporal}
\end{equation}
which, for pairs of a channel and a gradient, separates changes in the proposed motion, changes in the sensitivity of the task, and their joint change. These temporal terms enter the largest registry predictor, whereas the natural-history replay uses the symmetric attribution of Equations~\ref{eq:symmetricmotion}--\ref{eq:symmetricpolar}, and each convention has its own attribution terms.

For selected events, we evaluate the path integrals of Section~\ref{sec:finite}. The endpoint remainder is available directly from the endpoint losses and the initial slope, whereas resolving its components requires numerical integration. For the magnitude--orientation decomposition, the implementation records whether the cosine denominator exceeds $10^{-30}$ and divides by the maximum of the denominator and $10^{-30}$, so that exactly zero vectors yield zero, and every saved point on a population path in the inspected archives has a defined direction under this criterion.

We sample directional Hessian contractions, including $h^\top H_T h$, $h^\top H_T c$, $c^\top H_T h$, and $c^\top H_T c$, at locations where the slope varies rapidly, and these samples characterise the local curvature within the path. Layerwise contractions use full-model Hessian-vector products and therefore retain the couplings between layers.

\paragraph{Numerical checks and selection.}
Adaptive Simpson integration checks endpoint closure, agreement between nested grids, and the convergence of each channel and magnitude--orientation component. Directional slopes are checked by finite differences at the endpoints and the midpoint, Hessian-vector products receive separate finite-difference and mixed-symmetry checks, and acceptance requires the component checks as well as endpoint closure. The OLMo events were selected retrospectively from preliminary measurements of the trajectories, while for each other checkpoint, seed 1 contributes the update with the largest increase in validation loss and the update whose change in validation loss is closest to zero. Predictor fitting follows the separate source-only protocol below, and Appendix~\ref{app:audits} separates the resolved measurements from the failed checks.

\subsection{Prediction timing and persistence controls}
For the registry suite, logistic predictors fitted on OLMo are transferred with frozen coefficients and normalisation to the other checkpoints, with features ranging from losses to projections and recent geometry. For the text comparisons, a decision before update $d$ uses only records from update $d-1$ or earlier, so that the projection of the preceding update and the latest increase in validation loss are used to predict the next test-loss event. Rankings keep their fixed sign, and no score below chance is flipped. We distinguish same-update diagnostics from this forecast and compare the geometric scores with both fitted and simple persistence baselines, and Appendix~\ref{app:movedprediction} specifies the feature sets, thresholds, eligibility, and the separation between source and target.

\paragraph{Supporting activation interventions.}
Selected OLMo and replication states also receive fitted orthogonal activation patches, evaluated on held-out entities, with sham, inverse, mean-shift, scaling, and matched random controls. These interventions measure functional sensitivity while preserving the intrinsic Euclidean geometry, which tests how the selected states respond to a change in orientation, and Appendix~\ref{app:movedframe} gives the full fitting and factorial protocol.

\section{Supporting temporal, intervention, and retention comparisons}
\subsection{Local geometry is informative, while forecasting follows temporal persistence}
\label{sec:timingresults}
The frozen registry predictor improves over its fitted loss baseline on every target checkpoint (Table~\ref{tab:prediction_main}), and the text experiments then compare the geometry of the previous update with the latest available change in validation loss at the same decision time. Both signals forecast deterioration, but the latest loss change ranks events more accurately in each text seed (Table~\ref{tab:text_timing}), which identifies recently observed damage as a particularly effective summary of the information that persists into the next update.

The optimiser comparison makes this temporal dependence clearer, since on Pythia the projections remain strong diagnostics of the update being taken under all tested optimisers, but their forecasting value changes sharply when the projection of the preceding update is used instead, with Adam and momentum SGD keeping strong rankings while SGD and Adam without first-moment memory rank below chance in the fixed orientation of the score (Table~\ref{tab:optimizer_timing}). Loss increments likewise tend to be positively correlated across adjacent steps with first-moment memory and negatively correlated without it, a pattern that is consistent with optimiser-dependent persistence across the calibrated configurations.

\begin{table}[t]\centering\small
\caption{Local diagnostics versus advance forecasting in the text optimiser study. Entries are equal-weight means of seed-level AUROC for an $A$-test increase above 0.05 nats.}\label{tab:optimizer_timing}
\setlength{\tabcolsep}{4pt}
\begin{tabular}{llrrrr}\toprule
Model & Optimiser & $n$ & Same-update & Previous proj. & Latest loss\\\midrule
Pythia-410M & SGD & 3 & 0.979 & 0.386 & 0.388\\
Pythia-410M & Momentum SGD & 3 & 0.971 & 0.860 & 0.879\\
Pythia-410M & Adam $\beta_1=0$ & 3 & 0.966 & 0.433 & 0.417\\
Pythia-410M & Adam & 3 & 0.987 & 0.919 & 0.921\\
SmolLM2-135M & SGD & 3 & 0.867 & 0.436 & 0.446\\
SmolLM2-135M & Momentum SGD & 2 & 0.900 & 0.790 & 0.855\\
SmolLM2-135M & Adam $\beta_1=0$ & 3 & 0.659 & 0.561 & 0.593\\
SmolLM2-135M & Adam & 3 & 0.984 & 0.938 & 0.943\\
\bottomrule\end{tabular}
\smallskip\parbox{.97\linewidth}{\small Same-update projections measure local diagnostic performance, while preceding-update scores measure advance forecasting and have different eligible sample counts. The previous projection and the latest loss use the same preceding-update timestamp, and only acquisition-passed seeds enter these means, while the excluded momentum seed remains in Table~\ref{tab:optimizer_all}.}
\end{table}

In every acquisition-passed confirmation trajectory trained with Adam or momentum SGD, the latest increase in validation loss outperforms the preceding projection, so a practical warning signal is already present in the recent trace of the loss. SmolLM2 varies more in the configurations without first-moment memory, which include trajectories with sparse large events, and Appendix~\ref{app:optimizer} reports this dependence on the seed. The below-chance rankings keep their predeclared orientation and reflect how the direction of the temporal association changes across configurations. In the separate matched-start study with seeds 41--43, the frozen combined-feature predictor improves over its frozen multivariate loss-history baseline in all eight acquisition-passed trajectories, with a mean AUROC difference of $+0.074943$ at the next update and the 0.05-nat threshold, but the simple score given by the latest increase in validation loss exceeds the combined model in seven of the eight trajectories, so this simpler persistence score remains a strong comparator (Appendix~\ref{app:followup}).

\subsection{History interventions change retention beyond displacement size}
\label{sec:historycontrols}
Paired resets show that changing the first-moment history can alter retention while the size of the displacement is preserved. In the highlighted SmolLM2 momentum-SGD probe, the preserved update damages the old task while a block-matched reset improves both tasks, whereas an Adam probe shows the opposite dependence, with preserved history producing the greater improvement on both tasks (Table~\ref{tab:history_control} in the main text). The benefit of a reset therefore depends on how the stored history aligns with the sensitivity of the task at the chosen state, and the momentum-SGD result extends this observation to an optimiser with a non-adaptive learning rate.

Selected Pythia continuations follow these effects beyond the first displacement, since in one branch a block-matched reset reduces the immediate damage to the old task and finishes with lower losses on both tasks, while at another site the same intervention turns immediate improvements into increases, followed by a trade-off between retention and acquisition during the continuation (Table~\ref{tab:text_branch_details}). Because the matching applies only to the initial displacement, after which each branch evolves with its own optimiser state, the probes of the optimiser study and these short continuations measure, respectively, the immediate effects and some of their downstream consequences at selected sites.

The matched-start study of Appendix~\ref{app:followup} extends this comparison to 32-update branches, and averaged over three forks per acquired Adam trajectory, a block-matched reset at the first step improves the final old-task loss in all six model and seed trajectories, with mean effects of $-0.127$ nats for Pythia and $-0.267$ for SmolLM2, although the outcomes of individual forks and of the comparisons matched on acquisition are heterogeneous. For Pythia, the mean benefit combines worse confusion ($+0.100$) with improved leakage ($-0.227$), which links the outcome of the intervention to the functional decomposition, without equating it with the natural age attribution (Appendix~\ref{app:followup}).

The full optimiser trajectories also show substantial forgetting under plain SGD (Table~\ref{tab:optimizer_all}), and together with the reset probes, this establishes a state-dependent role for history within a broader range of forgetting dynamics.

\subsection{Forgetting usually begins in a locally unfavourable direction}

\begin{table}[t]\centering\small
\caption{Retention and local events. Loss columns average three seeds and event counts sum their 384 updates, and reversals require both $\Delta L_A>0.05$ and a negative initial slope.}\label{tab:retention_main}
\setlength{\tabcolsep}{4pt}
\begin{tabular}{lrrrr}\toprule
Checkpoint & Final $L_A$ minus anchor & Final $L_B$ & Large increases & Reversals\\\midrule
OLMo-1B & +0.466 & 0.118 & 78 & 2\\
Pythia-70M & +0.338 & 0.248 & 32 & 0\\
Pythia-410M & +0.787 & 0.136 & 76 & 1\\
Qwen2.5 base & +0.824 & 0.088 & 75 & 0\\
Qwen2.5 Instruct & +0.726 & 0.093 & 64 & 1\\
SmolLM2 base & +0.809 & 0.012 & 24 & 0\\
SmolLM2 Instruct & +0.745 & 0.031 & 19 & 0\\
Qwen3 base & +0.576 & 0.241 & 53 & 0\\
\bottomrule\end{tabular}\end{table}

Every registry trajectory ends with a higher old-task loss than at the post-acquisition anchor (Table~\ref{tab:retention_main}), and the deterioration also affects the answers themselves, since Qwen3 acquires the old task with perfect two-answer accuracy and then loses accuracy in every seed during new-task training (Table~\ref{tab:writing_retention_details}). Forgetting is therefore visible both in the probability assigned to the learned answers and in which answer the model selects.

The way damage accumulates differs across checkpoints, and SmolLM2, for instance, reaches a substantial loss of retention with relatively few large local events, which shows that event counts and final retention summarise different aspects of a trajectory. Comparisons also depend on how well the new task is acquired, since the Instruct variants have smaller mean losses of retention than their base counterparts, with mixed orderings at the level of seeds, and the smaller mean loss of retention of Qwen3 relative to Qwen2.5 base comes with weaker final performance on the new task. Table~\ref{tab:retention_main} therefore reports the two task outcomes together, and Appendix~\ref{app:events} gives the breakdown by seed.

In the registry trajectories, most large increases begin with an unfavourable old-task projection, while a small subset crosses from an initially favourable direction to a harmful endpoint as the sensitivity of the task evolves. The event counts in Table~\ref{tab:retention_main} and the breakdown by seed in Table~\ref{tab:retention_seeds} establish the prevalence of these events within this suite, and the selected path audits in the text and optimiser studies then examine how individual displacements produce their outcomes.

\paragraph{Selected finite-step exceptions.}
The OLMo event at seed 1/update 21 begins with favourable history and current projections but ends with a higher old-task loss, with an integrated history contribution of $+0.309796$ nats and a current contribution of $-0.021920$. Appendix~\ref{app:movedpath} gives the audited path and its magnitude--orientation decomposition, and Table~\ref{tab:retention_main} places this selected exception alongside the low prevalence of reversals in the full registry.

\paragraph{Channel balances differ across events.}
The other selected audited paths are dominated by history, dominated by the current gradient, or receive comparable contributions from both channels. In a selected Qwen3 registry event, both task losses rise because harmful integrated history outweighs a helpful current contribution on the new task, while another small event has protective history, and the full channel table is given in Appendix~\ref{app:movedchannels}. These paths establish that either channel can be harmful or protective at a selected state.

\subsection{Step-reduction policies produce variable trade-offs between retention and acquisition}
Using the projection to trigger a reduction of the step size has mixed effects on learning, since in one of the two Pythia text seeds the policy improves both final losses relative to natural training, while in the other both worsen, and its final old-task loss lies within the range of random schedules with matched action counts in each seed. Triggers based on the loss history and a lower constant learning rate remain competitive (Table~\ref{tab:text_policies}; Appendix~\ref{app:extensions}), so the timing of the intervention and its cost to new-task acquisition are central to evaluating a policy, alongside the quality of the underlying ranking signal.

\subsection{Probability leakage can dominate the loss of retained behaviour}
The answer-codebook controls reveal a striking gap between relative discrimination and the answers that the model actually produces. In the highlighted disjoint-label condition, the model still selects the correct old answer for most examples when restricted to the old answer set, yet its full-vocabulary accuracy falls to zero, and answer-mass leakage dominates its old-task loss (Table~\ref{tab:label_control}), so the model often ranks the correct old answer above its designated alternative while placing more probability elsewhere in the vocabulary.

Shared and disjoint codebooks start from identical old-task states within each seed, but their final retention differs, with disjoint labels increasing the old-task loss in some seeds and reducing it in another. Since changing the codebook of the new task also changes its instructions, the paired outcomes reflect both changes, and reporting conditional confusion, answer mass, and full-vocabulary accuracy makes these consequences visible even when restricted-answer discrimination remains useful.

\subsection{Activation interventions delimit the frame interpretation}
A fitted activation rotation restores two-answer accuracy in a selected OLMo event while preserving the norms, the covariance spectrum, and the distance-based persistence of the activations, and the fitted map helps the history-only and full-update states while harming the pre-update state. Across other OLMo sites and in the Pythia replication, the benefits are smaller, variable, or absent, so the functional response to the orientation of the activations depends on the selected state and on the fitted map. Appendix~\ref{app:movedrescue} gives the full outcomes and controls.

\section{Natural outcomes and history by gradient age}
\label{app:newreplication}

\begin{table}[htbp]\centering\small
\caption{Full-test accuracy before and after new-task training for Pythia and SmolLM2, in percent. Both accuracy measures share the same acquired anchor in these runs.}\label{tab:finalaccuracy}
\begin{tabular}{llrrr}\toprule
Model & Seed / answers & Anchor (both) & Final full & Final restricted\\\midrule
Pythia-410M & 51 / disjoint & 87.50 & 0.00 & 63.28\\
Pythia-410M & 51 / shared & 87.50 & 57.03 & 59.38\\
Pythia-410M & 52 / disjoint & 84.38 & 0.00 & 78.12\\
Pythia-410M & 52 / shared & 84.38 & 39.84 & 46.09\\
Pythia-410M & 53 / disjoint & 86.72 & 0.00 & 81.25\\
Pythia-410M & 53 / shared & 86.72 & 32.03 & 80.47\\
SmolLM2-135M & 51 / disjoint & 89.06 & 82.03 & 87.50\\
SmolLM2-135M & 51 / shared & 89.06 & 78.91 & 78.91\\
SmolLM2-135M & 52 / disjoint & 86.72 & 17.97 & 88.28\\
SmolLM2-135M & 52 / shared & 86.72 & 70.31 & 85.94\\
SmolLM2-135M & 53 / disjoint & 91.41 & 21.88 & 78.91\\
SmolLM2-135M & 53 / shared & 91.41 & 68.75 & 78.12\\
\bottomrule\end{tabular}\end{table}

\begin{table}[htbp]\centering\small
\caption{Cumulative leakage projections by gradient age on the fixed 32-text test subset, in nats, summed over 128 updates. The contributions keep their original magnitudes, bin widths, and decay weights under the current preconditioner, and the recorded roundoff in the sums is below $1.4\times10^{-7}$ for Pythia and SmolLM2 and below $10^{-6}$ for Qwen3.}\label{tab:historyages}\label{tab:qwenage}
\setlength{\tabcolsep}{4pt}
\begin{tabular}{llrrrr}\toprule
Model & Seed / answers & Task $A$ & B ages 1--4 & B ages 5--16 & B ages 17+\\\midrule
Pythia-410M & 51 / disjoint & +0.017229 & -0.649152 & +3.381657 & +1.840978\\
Pythia-410M & 51 / shared & +0.002387 & -0.374941 & +0.350112 & +0.198164\\
Pythia-410M & 52 / shared & +0.021419 & -1.845042 & +2.274017 & +0.217016\\
SmolLM2-135M & 51 / disjoint & -0.000818 & -0.251938 & +0.347463 & +0.182692\\
SmolLM2-135M & 51 / shared & -0.005015 & -0.106745 & +0.152827 & +0.037062\\
SmolLM2-135M & 52 / disjoint & +0.003544 & +0.030012 & +1.183401 & +0.641751\\
SmolLM2-135M & 52 / shared & +0.002240 & -0.264665 & +0.474573 & +0.092687\\
\midrule
Qwen3-0.6B & 61 / disjoint & +0.128180 & -3.374246 & +12.291977 & +2.136861\\
Qwen3-0.6B & 61 / shared & -0.002562 & -1.211178 & +3.793636 & +0.752107\\
Qwen3-0.6B & 62 / disjoint & +0.097479 & -2.350902 & +5.733983 & +1.788376\\
Qwen3-0.6B & 62 / shared & -0.003009 & -2.217560 & +4.211019 & +0.335133\\
Qwen3-0.6B & 63 / disjoint & -0.037396 & -0.226403 & +5.418361 & +1.861431\\
Qwen3-0.6B & 63 / shared & -0.007850 & -1.165692 & +2.867342 & +0.556583\\
\bottomrule\end{tabular}\end{table}

The natural endpoint populations contain 128 fixed test examples per task, with Pythia and SmolLM2 each contributing seeds 51--53 under the two paired answer conditions and Qwen contributing seeds 61--63 under the same conditions, and every trajectory passes acquisition and completes 128 task-$B$ updates. Table~\ref{tab:newendpoints} reports the changes relative to the acquired anchor of each trajectory. The 18 trajectories together contain 2,304 new-task updates, although they are not 18 independent seeds, and these endpoints do not replace the replay outcomes on 32 examples in Table~\ref{tab:historyleakage}.

\begin{table}[htbp]\centering\small
\caption{Endpoint changes on the full 128-text test set after 128 updates, where D/S denote disjoint and shared answers. All trajectories pass acquisition.}\label{tab:newendpoints}
\begin{tabular}{llrrr}\toprule
Model & Seed / answers & Total & Confusion & Leakage\\\midrule
Pythia-410M & 51 / disjoint & +4.703567 & -0.099648 & +4.803215\\
Pythia-410M & 51 / shared & +1.269549 & +1.164286 & +0.105263\\
Pythia-410M & 52 / disjoint & +4.027747 & -0.830915 & +4.858662\\
Pythia-410M & 52 / shared & +0.297949 & +0.051880 & +0.246069\\
Pythia-410M & 53 / disjoint & +6.158765 & -0.285333 & +6.444098\\
Pythia-410M & 53 / shared & +1.418819 & -0.185479 & +1.604298\\
SmolLM2-135M & 51 / disjoint & -0.171659 & -0.416108 & +0.244449\\
SmolLM2-135M & 51 / shared & -0.179211 & -0.211255 & +0.032044\\
SmolLM2-135M & 52 / disjoint & +1.543555 & -0.394955 & +1.938509\\
SmolLM2-135M & 52 / shared & +0.079588 & -0.374151 & +0.453740\\
SmolLM2-135M & 53 / disjoint & +1.861977 & -0.010326 & +1.872303\\
SmolLM2-135M & 53 / shared & +0.278116 & -0.031183 & +0.309299\\
Qwen3-0.6B & 61 / disjoint & +9.796397 & -0.526192 & +10.322589\\
Qwen3-0.6B & 61 / shared & +2.336601 & -0.666572 & +3.003173\\
Qwen3-0.6B & 62 / disjoint & +3.788080 & -0.444776 & +4.232856\\
Qwen3-0.6B & 62 / shared & +1.141700 & -0.261808 & +1.403508\\
Qwen3-0.6B & 63 / disjoint & +6.377646 & -0.478961 & +6.856607\\
Qwen3-0.6B & 63 / shared & +1.258086 & -0.529663 & +1.787748\\
\bottomrule\end{tabular}\end{table}

Qwen uses the pinned Qwen3-0.6B-Base revision already listed for the registry suite, with FP32 native Adam, eager attention, dropout disabled, 600 task-$A$ updates, and a learning rate of $2\times10^{-5}$. Its six history replays complete 768 updates, with all endpoint hashes, algebra checks, and age audits passing and a largest relative residual of $1.3283\times10^{-7}$ in the reconstruction of the moment, and the component gradients and matched outcomes use all 32 validation texts and the first 32 test texts in fixed order. Together with the seven Pythia and SmolLM2 replays, which cover 896 updates, the evidence on gradient age totals 13 trajectories and 1,664 updates, and the finite-difference validity along interpolated paths is assessed through the separate derivative checks. Table~\ref{tab:qwenage} reports the Qwen age decomposition alongside the Pythia and SmolLM2 results.

\begin{table}[htbp]\centering\small
\caption{Absolute-contribution shares, in percent, for changes in the total-loss history projection of Qwen. Orientation, channel motion, and denominator change use the separate or nested normalisations of Appendix~\ref{app:historydynamics}, so the columns are not additive.}\label{tab:qwenmotion}
\begin{tabular}{llrrr}\toprule
Seed & Answers & Orientation & Channel motion & Denominator change\\\midrule
61 & disjoint & 78.18 & 64.19 & 3.17\\
61 & shared & 83.34 & 73.23 & 2.63\\
62 & disjoint & 77.89 & 59.10 & 3.23\\
62 & shared & 77.93 & 69.92 & 2.22\\
63 & disjoint & 83.97 & 73.56 & 2.85\\
63 & shared & 78.25 & 69.78 & 1.84\\
\bottomrule\end{tabular}\end{table}

Qwen's exported cohort records follow task-$B$ gradients originating at updates 1, 8, 32, and 64 through their subsequent ages 1--32, and they retain both the current adaptive denominator and a comparison with the denominator fixed at the origin, under the current clock and decay. These diagnostics track fixed gradient contributions as the state of the model evolves, whereas the aggregate bins summarise their measured effects across the recorded trajectories.

\section{Finite-path sampling and numerical validation}
\label{app:newpaths}
The Pythia and SmolLM2 finite-path study uses 12 natural trajectories, 1,536 new-task updates, and 252 validation and test population paths, and all trajectories pass acquisition. Every component audit passes for 204 of these paths, namely 82 of the 130 Pythia paths and all 122 SmolLM2 paths, while the remaining 48 Pythia paths have at least one unresolved component. Considering each component separately, total loss passes on 228 paths (116 validation and 112 test), confusion on 217, and leakage on 234, and population paths at the same update share their parameters and are therefore dependent.

All 96 planned uniform paths are completed in each population, and among the uniform test paths, 87 are accepted for total loss, 81 for confusion, and 89 for leakage, with the corresponding unresolved counts listed in Table~\ref{tab:coverage}. The full selected pool also contains enriched events, and Table~\ref{tab:finalcounts} reports the two selection scopes separately. Coupling frequencies describe the accepted paths under the stated thresholds, and the coupling of unresolved paths is recorded as unknown. The endpoint and finite-path analyses use all twelve Pythia and SmolLM2 trajectories, whereas the history-age analysis uses seven of these trajectories together with six Qwen trajectories.

\begin{table}[htbp]\centering\small
\caption{Path counts for each component. N: completed; Acc.: accepted by the audit; U: unresolved; Small: ordinary or nonlarge; FO: first-order dominated; Amp.: nonlinear amplification; Mix: material mixed term. The three event columns partition the accepted paths, since no accepted path is classified as a reversal or as another large increase, while Mix is an independent flag. Uniform rows are subsets of the corresponding rows for all selected paths.}\label{tab:finalcounts}
\setlength{\tabcolsep}{3.5pt}
\begin{tabular}{lllrrrrrrr}\toprule
Population & Selection & Component & N & Acc. & U & Small & FO & Amp. & Mix\\\midrule
Validation & All & Total & 126 & 116 & 10 & 85 & 29 & 2 & 1\\
Validation & All & Confusion & 126 & 113 & 13 & 108 & 5 & 0 & 1\\
Validation & All & Leakage & 126 & 118 & 8 & 93 & 23 & 2 & 0\\
Validation & Uniform & Total & 96 & 90 & 6 & 74 & 14 & 2 & 1\\
Validation & Uniform & Confusion & 96 & 88 & 8 & 88 & 0 & 0 & 1\\
Validation & Uniform & Leakage & 96 & 91 & 5 & 75 & 14 & 2 & 0\\
Test & All & Total & 126 & 112 & 14 & 80 & 30 & 2 & 1\\
Test & All & Confusion & 126 & 104 & 22 & 95 & 9 & 0 & 1\\
Test & All & Leakage & 126 & 116 & 10 & 92 & 22 & 2 & 0\\
Test & Uniform & Total & 96 & 87 & 9 & 68 & 17 & 2 & 1\\
Test & Uniform & Confusion & 96 & 81 & 15 & 76 & 5 & 0 & 1\\
Test & Uniform & Leakage & 96 & 89 & 7 & 74 & 13 & 2 & 0\\
\bottomrule\end{tabular}\end{table}

\subsection{Curvature of the paired amplification paths}

\begin{table}[htbp]\centering\small
\caption{Paired uniform paths for Pythia seed 53/update 9 on the 32-text test subset, all passing every component audit. Positive remainders coexist with protective mixed terms, and values are rounded to six decimals.}\label{tab:newamplification}
\begin{tabular}{lrrrr}\toprule
Answers & Observed & Initial projection & Remainder & Mixed term\\\midrule
Disjoint & +0.099559 & +0.049355 & +0.050205 & -0.006253\\
Shared & +0.139375 & +0.098201 & +0.041175 & -0.054475\\
\bottomrule\end{tabular}\end{table}

\begin{table}[htbp]\centering\small
\caption{Weighted curvature integrals for the fully audited paired amplification examples, where $hh$ and $cc$ denote the diagonal directional terms, $hc+ch$ their mixed contribution, and $dd$ the integral over the full displacement, with closure defined as $dd-R$. Values are in nats, and the small residuals in the direction sums are retained in the export.}\label{tab:finalcurvature}
\setlength{\tabcolsep}{4pt}
\begin{tabular}{lrrrrr}\toprule
Answers & $hh$ & $cc$ & $hc+ch$ & $dd$ & Closure\\\midrule
Disjoint & +0.055557 & +0.000814 & -0.006253 & +0.050176 & -0.000028\\
Shared & +0.082203 & +0.013482 & -0.054475 & +0.041202 & +0.000028\\
\bottomrule\end{tabular}\end{table}

The paired test paths for Pythia seed 53/update 9 pass all component audits, and Table~\ref{tab:newamplification} gives their observed changes and initial projections, while Table~\ref{tab:finalcurvature} resolves the weighted curvature integrals. In both cases, the history--history term supplies the largest positive contribution to the curvature and the mixed terms are negative. These terms describe fixed directions along the actual full displacement, whereas an intervention that trained with history alone would induce a different path.

The residuals in the direction sums are $+5.810694\times10^{-5}$ for disjoint answers and $-7.868590\times10^{-6}$ for shared answers, and the individual recorded $hr$, $cr$, and $rr$ terms have magnitudes below $10^{-8}$. The one material mixed term for total loss in the accepted uniform test sample is negative, and no accepted uniform leakage path meets the materiality threshold for the mixed term. These observations are subject to the model-dependent numerical acceptance described above.

\subsection{Functional accuracy and paired codebooks}
Table~\ref{tab:newendpoints} reports the endpoint losses on the full test set, and Table~\ref{tab:finalaccuracy} gives the full-vocabulary and restricted-answer accuracy on the same 128 examples. Leakage rises in every Pythia and SmolLM2 trajectory, while confusion decreases in ten cases and total loss in two, and full-vocabulary accuracy falls throughout. Each paired comparison verifies that the acquired old-task weights and the selected examples are identical, and disjoint codebooks produce greater leakage and greater deterioration of total loss in all six pairs, as well as higher restricted accuracy in all six, while their full-vocabulary accuracy is lower in five. The codebook, instructions, and token identities of the new task change together between the two conditions.

\subsection{Operational event classification}
\label{app:classification}
For a population path, we set $\Delta=L(\theta+\delta)-L(\theta)$, $P=\nabla L(\theta)^\top\delta$, and $R=\Delta-P$. The classifier applies the following ordered rules and stops at the first match:
\begin{enumerate}
\item Failed integration or derivative audits for the evaluated loss component: \texttt{unresolved}.
\item $\Delta\leq0.05$ nats: \texttt{ordinary\_or\_nonlarge}.
\item $P<-0.01$ nats: \texttt{favorable\_to\_harmful\_reversal}.
\item $R>0.01$ nats and $R\geq0.25|\Delta|$: \texttt{nonlinear\_amplification}.
\item $P>0$ and $|P|\geq|R|$: \texttt{first\_order\_dominated}.
\item Otherwise: \texttt{other\_large\_increase}.
\end{enumerate}
The amplification rule takes precedence over first-order dominance, so that, for example, $P=0.07$, $R=0.03$, and $\Delta=0.10$ counts as amplification, because the remainder supplies 30\% of the observed increase, and amplification refers to the entire remainder. The separate \texttt{mixed\_material} flag requires $|M|\geq0.01$ nats and $|M|\geq0.25|\Delta|$, regardless of the sign of the mixed term. These are operational thresholds rather than significance tests or privileged mathematical boundaries, and the reversal threshold $P<-0.01$ used for the natural components differs from the threshold $P<0$ used in the registry study.

Audit acceptance and classification are evaluated separately for total loss, confusion, and leakage. All 756 component classifications for Pythia and SmolLM2 reproduce this ordered rule, and the same holds for all 72 Qwen3 component records. Acceptance of a full path additionally requires every component to pass, whereas the counts for each component use the gate of that component, and the mixed-material flag is independent of the classification of large events, so it can be set on an ordinary or nonlarge path.

\clearpage
\subsection{Qwen3 finite-path measurements}
\label{app:qwenfinalpaths}

\begin{table}[htbp]\centering\small
\caption{Qwen3 held-out paths for total loss. D/S: disjoint/shared answers; U/L: uniform or validation-selected large-$A$ case. $\Delta$ is the observed change, $P$ the initial projection, $R=\Delta-P$, and $M$ the mixed contribution, all in nats and rounded to six decimals, and all component audits pass.}\label{tab:qwenpathtotal}
\setlength{\tabcolsep}{4pt}
\begin{tabular}{llrlrrrr}\toprule
Seed & Answers & Step & Select. & $\Delta$ & $P$ & $R$ & $M$\\\midrule
61 & D & 73 & U & -0.036533 & -0.062171 & +0.025638 & +0.005214\\
61 & D & 103 & L & +0.066291 & +0.062903 & +0.003387 & +0.001178\\
61 & S & 73 & U & +0.070418 & +0.063665 & +0.006753 & -0.005559\\
61 & S & 94 & L & +0.129903 & +0.130472 & -0.000569 & -0.000319\\
62 & D & 51 & L & +0.176322 & +0.164273 & +0.012049 & +0.001512\\
62 & D & 121 & U & -0.074695 & -0.073804 & -0.000890 & +0.000330\\
62 & S & 87 & L & +0.159014 & +0.149593 & +0.009421 & -0.003291\\
62 & S & 121 & U & +0.088379 & +0.086205 & +0.002174 & -0.001492\\
63 & D & 12 & L & +0.196892 & +0.169961 & +0.026931 & +0.015216\\
63 & D & 51 & U & +0.001897 & +0.001490 & +0.000407 & -0.002354\\
63 & S & 2 & L & -0.009480 & -0.002950 & -0.006530 & -0.003967\\
63 & S & 51 & U & -0.056687 & -0.063346 & +0.006659 & -0.000356\\
\bottomrule\end{tabular}\end{table}

The Qwen3 finite-path study comprises six acquired trajectories, 768 updates, and 24 population paths, and its six history replays pass all endpoint-hash and algebra checks and all 768 age-reconstruction audits. These are the six replays reported in Appendix~\ref{app:newreplication}, and the finite-path sample described below is drawn from the same trajectories.

The path sample contains 12 distinct cases of trajectory and update, namely one uniformly selected update and one large event in old-task loss, selected on validation, per trajectory, and the shared and disjoint conditions use the same seeded uniform draw of the step. Each case is evaluated on fixed validation and test populations of 32 examples, which yields 24 population paths, and all component integration and sampled derivative audits pass, with all 72 component records satisfying the ordered classification and materiality rules. These numerical checks assess the sampled derivatives and integrals, while the paired conditions, the validation and test evaluations, and overlapping event windows remain dependent.

\begin{table}[htbp]\centering\small
\caption{Qwen3 component classifications by population and selection, where every row contains six completed and accepted paths. Small: ordinary or nonlarge; FO: first-order dominated. No reversal, amplification, other large increase, or material mixed term occurs, and the large-$A$ cases are selected on validation loss.}\label{tab:qwenfinalcounts}
\begin{tabular}{lllrr}\toprule
Population & Selection & Component & Small & FO\\\midrule
Validation & Uniform & Total & 4 & 2\\
Validation & Uniform & Confusion & 6 & 0\\
Validation & Uniform & Leakage & 4 & 2\\
Validation & Large-$A$ & Total & 0 & 6\\
Validation & Large-$A$ & Confusion & 4 & 2\\
Validation & Large-$A$ & Leakage & 1 & 5\\
Test & Uniform & Total & 4 & 2\\
Test & Uniform & Confusion & 6 & 0\\
Test & Uniform & Leakage & 4 & 2\\
Test & Large-$A$ & Total & 1 & 5\\
Test & Large-$A$ & Confusion & 6 & 0\\
Test & Large-$A$ & Leakage & 1 & 5\\
\bottomrule\end{tabular}\end{table}

The uniform test sample contains two large increases in total loss and the validation-selected sample contains five, all seven of which are first-order dominated and also show large increases in leakage, while none of the selected test paths has a large increase in confusion. One large event selected on validation improves the total test loss, which illustrates why held-out outcomes are evaluated on their own population (Table~\ref{tab:qwenfinalcounts}). These small samples, selected in different ways, describe the measured paths without estimating a common prevalence across models.

Mixed curvature takes both signs, since at update 12 of seed 63 with disjoint answers, its contribution to total loss is positive and exceeds the absolute cutoff but supplies only about 7.7\% of the observed increase, which is below the relative materiality threshold, whereas at update 73 of seed 61 with shared answers, negative mixed curvature offsets positive history--history curvature, and the full remainder stays positive but falls below the amplification cutoff. First-order dominance can therefore coexist with nonlinear terms that either reinforce or protect, and Tables~\ref{tab:qwenpathtotal}, \ref{tab:qwenpathconfusion}, and~\ref{tab:qwenpathleakage} give their signed values.

\begin{table}[htbp]\centering\small
\caption{Qwen3 held-out paths for confusion loss. D/S: disjoint/shared answers; U/L: uniform or validation-selected large-$A$ case. $\Delta$ is the observed change, $P$ the initial projection, $R=\Delta-P$, and $M$ the mixed contribution, all in nats and rounded to six decimals, and all component audits pass.}\label{tab:qwenpathconfusion}
\setlength{\tabcolsep}{4pt}
\begin{tabular}{llrlrrrr}\toprule
Seed & Answers & Step & Select. & $\Delta$ & $P$ & $R$ & $M$\\\midrule
61 & D & 73 & U & -0.030631 & -0.028159 & -0.002472 & -0.000566\\
61 & D & 103 & L & -0.003343 & -0.003699 & +0.000356 & +0.000086\\
61 & S & 73 & U & +0.015562 & +0.015279 & +0.000283 & -0.000209\\
61 & S & 94 & L & -0.026391 & -0.027637 & +0.001247 & +0.000602\\
62 & D & 51 & L & -0.027878 & -0.028722 & +0.000843 & +0.000314\\
62 & D & 121 & U & +0.012740 & +0.011925 & +0.000816 & -0.001797\\
62 & S & 87 & L & +0.044828 & +0.039387 & +0.005441 & +0.001683\\
62 & S & 121 & U & -0.009258 & -0.009305 & +0.000048 & -0.000818\\
63 & D & 12 & L & -0.028751 & -0.018684 & -0.010066 & -0.002073\\
63 & D & 51 & U & +0.006410 & +0.006454 & -0.000044 & -0.000464\\
63 & S & 2 & L & -0.009492 & -0.002960 & -0.006532 & -0.003967\\
63 & S & 51 & U & -0.025873 & -0.030027 & +0.004153 & -0.000272\\
\bottomrule\end{tabular}\end{table}

\begin{table}[htbp]\centering\small
\caption{Qwen3 held-out paths for leakage loss. D/S: disjoint/shared answers; U/L: uniform or validation-selected large-$A$ case. $\Delta$ is the observed change, $P$ the initial projection, $R=\Delta-P$, and $M$ the mixed contribution, all in nats and rounded to six decimals, and all component audits pass.}\label{tab:qwenpathleakage}
\setlength{\tabcolsep}{4pt}
\begin{tabular}{llrlrrrr}\toprule
Seed & Answers & Step & Select. & $\Delta$ & $P$ & $R$ & $M$\\\midrule
61 & D & 73 & U & -0.005902 & -0.034012 & +0.028110 & +0.005780\\
61 & D & 103 & L & +0.069634 & +0.066602 & +0.003032 & +0.001093\\
61 & S & 73 & U & +0.054856 & +0.048386 & +0.006470 & -0.005350\\
61 & S & 94 & L & +0.156293 & +0.158109 & -0.001816 & -0.000921\\
62 & D & 51 & L & +0.204200 & +0.192994 & +0.011205 & +0.001198\\
62 & D & 121 & U & -0.087435 & -0.085729 & -0.001706 & +0.002127\\
62 & S & 87 & L & +0.114186 & +0.110206 & +0.003980 & -0.004974\\
62 & S & 121 & U & +0.097637 & +0.095510 & +0.002126 & -0.000674\\
63 & D & 12 & L & +0.225642 & +0.188646 & +0.036997 & +0.017289\\
63 & D & 51 & U & -0.004513 & -0.004964 & +0.000451 & -0.001890\\
63 & S & 2 & L & +0.000012 & +0.000009 & +0.000003 & +0.000001\\
63 & S & 51 & U & -0.030813 & -0.033319 & +0.002506 & -0.000085\\
\bottomrule\end{tabular}\end{table}

Across the 36 held-out component records, the largest absolute difference between the curvature integral over the full displacement and $R$ is $4.99\times10^{-6}$ nats, and the largest absolute recorded residual in the direction sums is $2.37\times10^{-9}$ nats. The paired amplification examples in Table~\ref{tab:finalcurvature} and the Qwen3 paths therefore document different balances between initial projections and finite-step terms under the same operational thresholds.

\clearpage
\section{Extended related work}
\label{app:extendedrelated}
Our analysis combines three established ideas, namely constraints based on task gradients, the memory of the optimiser, and path attribution, and it is their combination that lets us evaluate stored directions against specific components of retained behaviour.

\subsection{Retaining behaviour through parameter and functional constraints}
Elastic weight consolidation protects parameters according to their estimated importance for previously learned tasks \citep{kirkpatrick2017overcoming}, Synaptic Intelligence accumulates the contribution of each parameter to the improvement of the loss along a learning trajectory and uses the resulting importance estimates to regularise subsequent learning \citep{zenke2017synaptic}, and Learning without Forgetting preserves earlier predictions on the inputs of the new task through a distillation objective \citep{li2016learning}. These approaches connect retention to the sensitivity of parameters, to the history of learning, and to the input--output behaviour of the model, and our measurements bring these perspectives together at the scale of a single applied update, where a stored contribution of the optimiser receives a functional interpretation through its projection onto a specified component of old-task loss. In particular, the history channel contains gradients accumulated during optimisation, whereas an estimate of parameter importance records how movement contributed to an earlier objective, and resolving history by origin and age shows which stored gradients currently favour leakage and which oppose it.

\subsection{Gradient interference and the direction of the applied update}
Gradient Episodic Memory uses remembered examples to constrain updates through their inner products with the gradients of earlier tasks \citep{lopezpaz2017gradient}, Orthogonal Gradient Descent projects new-task gradients onto the subspace orthogonal to stored gradients of the model outputs \citep{farajtabar2020orthogonal}, and in multitask learning, PCGrad modifies conflicting task gradients by projecting away their opposing components \citep{yu2020gradient}. Together, these methods establish gradient alignment as a useful description of interference and as a basis for controlling learning. Our channel projections evaluate the signed displacement produced by the optimiser, including its stored first moment and its adaptive scaling, against the gradients of the total loss, of confusion, and of leakage, and this component-level view explains how one update can improve discrimination within the old answer set while reducing the probability assigned to that set. Finite-path integration then measures how these sensitivities evolve between the recorded starting and ending parameters, so that the initial projection and the integrated contribution answer the corresponding local and finite-step questions about the same displacement.

\subsection{Momentum, gradient age, and retention}
The related-work section of the main text discusses momentum filtering, gradient age, and task-relative alignment, including MoFO, AdEMAMix, cautious optimisers, and optimiser-dependent task interference.

\subsection{Loss geometry and finite-path attribution}
\citet{li2024revisiting} relate forgetting during the tuning of language models to the flatness of the loss landscape and investigate sharpness-aware minimisation, and work on the edge of stability documents how curvature and finite learning rates shape the optimisation of neural networks, including short-term nonmonotonic dynamics of the loss \citep{cohen2021edge}. Our finite-path measurements examine the curvature encountered along the direction of the update that was actually applied, and expanding the displacement into history, current, and residual channels separates the corresponding diagonal and mixed contributions to the Taylor remainder. The fully audited amplification examples show why the sign of each term matters, since positive history--history curvature can outweigh a protective mixed interaction, while the frequency analysis separately records the size of the full remainder and the materiality of the mixed term.

Path integration also has a direct mathematical connection to attribution methods, since Integrated Gradients attributes a difference in prediction to input coordinates by integrating derivatives along a path from a baseline \citep{sundararajan2017axiomatic}, and in our setting the path lies in parameter space, its endpoints are successive recorded iterates, and the attributed directions are optimiser channels. By linearity of the directional derivative, the channel integrals sum to the loss change whenever the decomposition of the displacement and the integration are exact, and numerical closure and derivative audits assess this identity in the measured model. This connection places the analysis within established path-attribution mathematics while identifying its experimental object, namely the finite displacement that is actually applied during sequential learning.

\subsection{Answer competition and the functional form of forgetting}
Class-incremental learning has long highlighted the competition between old and new outputs, with \citet{wu2019large} identifying a bias towards new classes in the final classifier and using a correction to improve retention, and in the adaptation of language models, the vocabulary supplies a larger set of competing outputs. Our confusion--leakage identity follows directly from factoring the probability of the correct token into its conditional probability within the designated answer set and the total mass of that set, and it provides an exact decomposition into two behaviours that can move in opposite directions. The paired codebooks make the competition between answers visible, while full-vocabulary and restricted-answer accuracy connect the loss components to answer selection, and projecting the gradient of each component onto the recorded optimiser channels then links this functional distinction to the dynamics that produce it.

\subsection{Representation change and functional sensitivity}
\citet{kumar2022finetuning} show that fine-tuning can distort pretrained features and impair out-of-distribution generalisation, with theory and experiments linking this outcome to the joint adaptation of the features and the prediction head, and methods for comparing representations depend on which transformations their similarity measures treat as equivalent \citep{kornblith2019similarity}. These observations motivate examining both the geometry of activations and its relationship to the downstream computation. Our supporting orthogonal activation interventions preserve pairwise Euclidean distances while changing the orientation presented to the remaining layers, and their state-dependent effects show how retained geometry and functional behaviour can vary independently under the chosen intervention. This complements the central analysis in parameter space, since in both cases the relevant direction is defined relative to the computation whose behaviour is being measured.
\end{document}